\documentclass[sigconf]{acmart}
\usepackage{hyperref}
\usepackage{algorithm}
\usepackage{algorithmic}
\usepackage{multirow}
\usepackage{booktabs}

\AtBeginDocument{%
  }

\setcopyright{acmcopyright}
\copyrightyear{2018}
\acmYear{2018}
\acmDOI{XXXXXXX.XXXXXXX}

\acmConference[Conference acronym 'XX]{Make sure to enter the correct
  conference title from your rights confirmation emai}{June 03--05,
  2018}{Woodstock, NY}
\acmPrice{15.00}
\acmISBN{978-1-4503-XXXX-X/18/06}

\acmSubmissionID{928}

\allowdisplaybreaks

\usepackage{subcaption}
\newcommand\restr[2]{{\left.\kern-\nulldelimiterspace{}#1\right|_{#2}}}

\newcommand{\TWORCell}[2]{\begin{tabular}{@{}c@{}}#1 \\ #2\end{tabular}}

\def\twowords#1 #2\relax{%
    \def\imgtrimwidth{#1}%
    \def\imgtrimheight{#2}%
}
\def\fourwords#1 #2 #3 #4\relax{
    \def\imgtrimleft{#1}%
    \def\imgtrimbottom{#2}%
    \def\imgtrimright{#3}%
    \def\imgtrimtop{#4}amatuer%
}

\usepackage{wrapfig}

\setcopyright{acmlicensed}
\acmYear{2023} 
\acmVolume{42} \acmNumber{4} 
\acmArticle{1}
\acmMonth{8} \acmPrice{15.00}\acmDOI{10.1145/3592143}

\definecolor{myblue}{RGB}{0, 82, 217}
\definecolor{myorange}{RGB}{238, 126, 71}

\begin{document}

\title{TexPainter: Generative Mesh Texturing with Multi-view Consistency}

\author{Hongkun Zhang}
\orcid{0000-0002-3199-5560}
\email{H.K.Zhang5813@gmail.com}
\affiliation{%
  \institution{Southeast University}
    \city{Nanjing}
  \state{Jiangsu}
  \country{China}
}

\author{Zherong Pan}
\orcid{0000-0001-9348-526X}
\email{zrpan@global.tencent.com}
\affiliation{%
  \institution{LightSpeed Studios}
    \city{Seattle}
  \state{WA}
  \country{USA}
}

\author{Congyi Zhang}
\orcid{0000-0002-4259-2863}
\email{congyiz@cs.ubc.ca}
\affiliation{%
    \institution{The University of British Columbia}
    \city{Vancouver}
    \state{BC}
    \country{Canada}
}
\affiliation{%
    \institution{TransGP \& HKU, Hong Kong}
    \city{}
    \country{}
}

\author{Lifeng Zhu}
\authornote{Corresponding author}
\orcid{0000-0002-9999-4513}
\email{lfzhulf@gmail.com}
\affiliation{%
  \institution{Southeast University}
    \city{Nanjing}
  \state{Jiangsu}
  \country{China}
}

\author{Xifeng Gao}
\orcid{0000-0003-0829-7075}
\email{xifgao@global.tencent.com}
\affiliation{%
  \institution{LightSpeed Studios}
    \city{Seattle}
  \state{WA}
  \country{USA}
}

\begin{abstract}
The recent success of pre-trained diffusion models unlocks the possibility of the automatic generation of textures for arbitrary 3D meshes in the wild. However, these models are trained in the screen space, while converting them to a multi-view consistent texture image poses a major obstacle to the output quality. In this paper, we propose a novel method to enforce multi-view consistency. Our method is based on the observation that latent space in a pre-trained diffusion model is noised separately for each camera view, making it difficult to achieve multi-view consistency by directly manipulating the latent codes. Based on the celebrated Denoising Diffusion Implicit Models (DDIM) scheme, we propose to use an optimization-based color-fusion to enforce consistency and indirectly modify the latent codes by gradient back-propagation. Our method further relaxes the sequential dependency assumption among the camera views. By evaluating on a series of general 3D models, we find our simple approach improves consistency and overall quality of the generated textures as compared to competing state-of-the-arts. Our implementation is available at: \url{https://github.com/Quantuman134/TexPainter}
\end{abstract}



\begin{teaserfigure}
\centering
\setlength{\tabcolsep}{0px}
\begin{tabular}{cccc}
\includegraphics[width=0.2\linewidth,trim=100px 100px 100px 100px,clip]{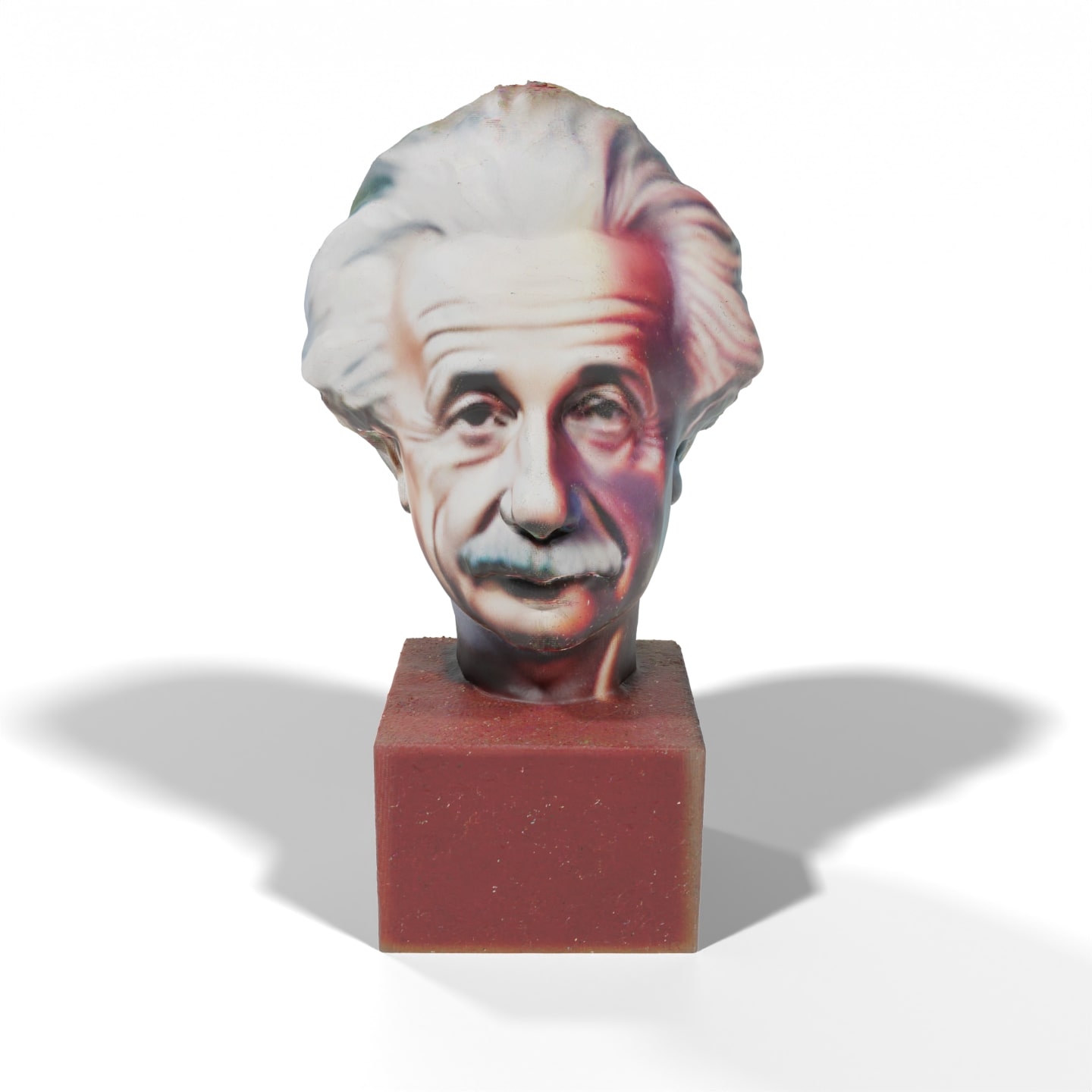}&
\includegraphics[width=0.2\linewidth,trim=100px 100px 100px 100px,clip]{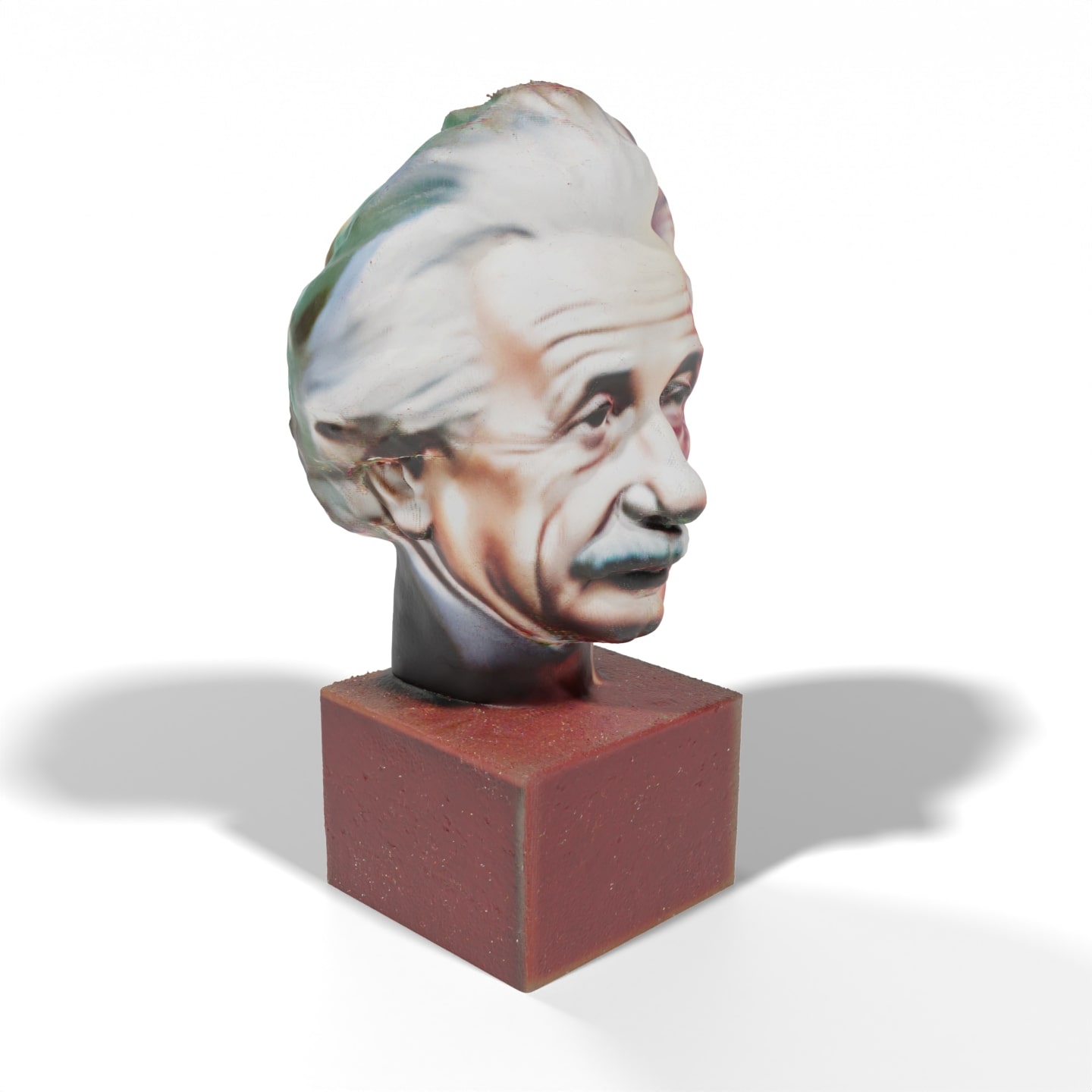}&
\includegraphics[width=0.2\linewidth,trim=250px 250px 250px 250px,clip]{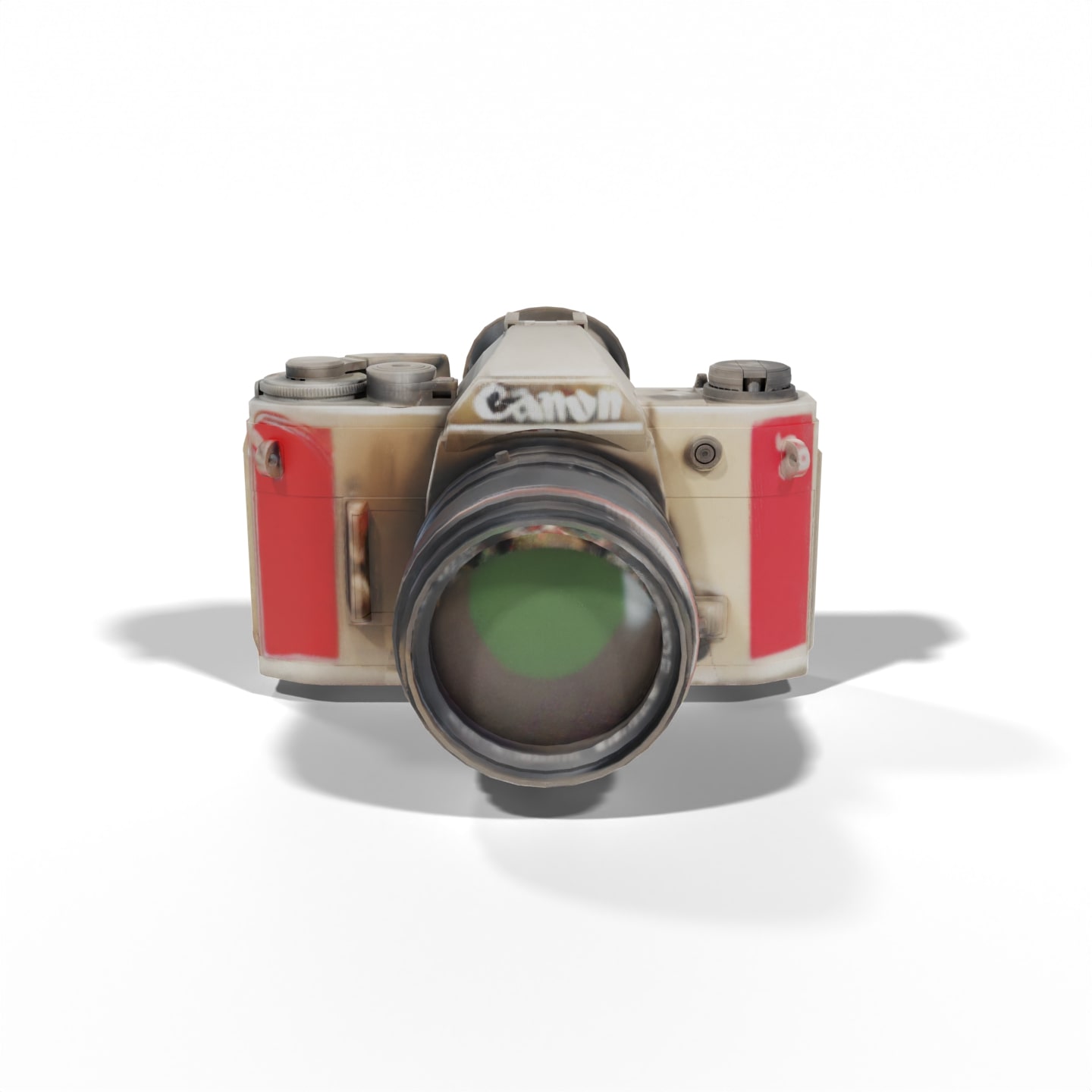}&
\includegraphics[width=0.2\linewidth,trim=250px 250px 250px 250px,clip]{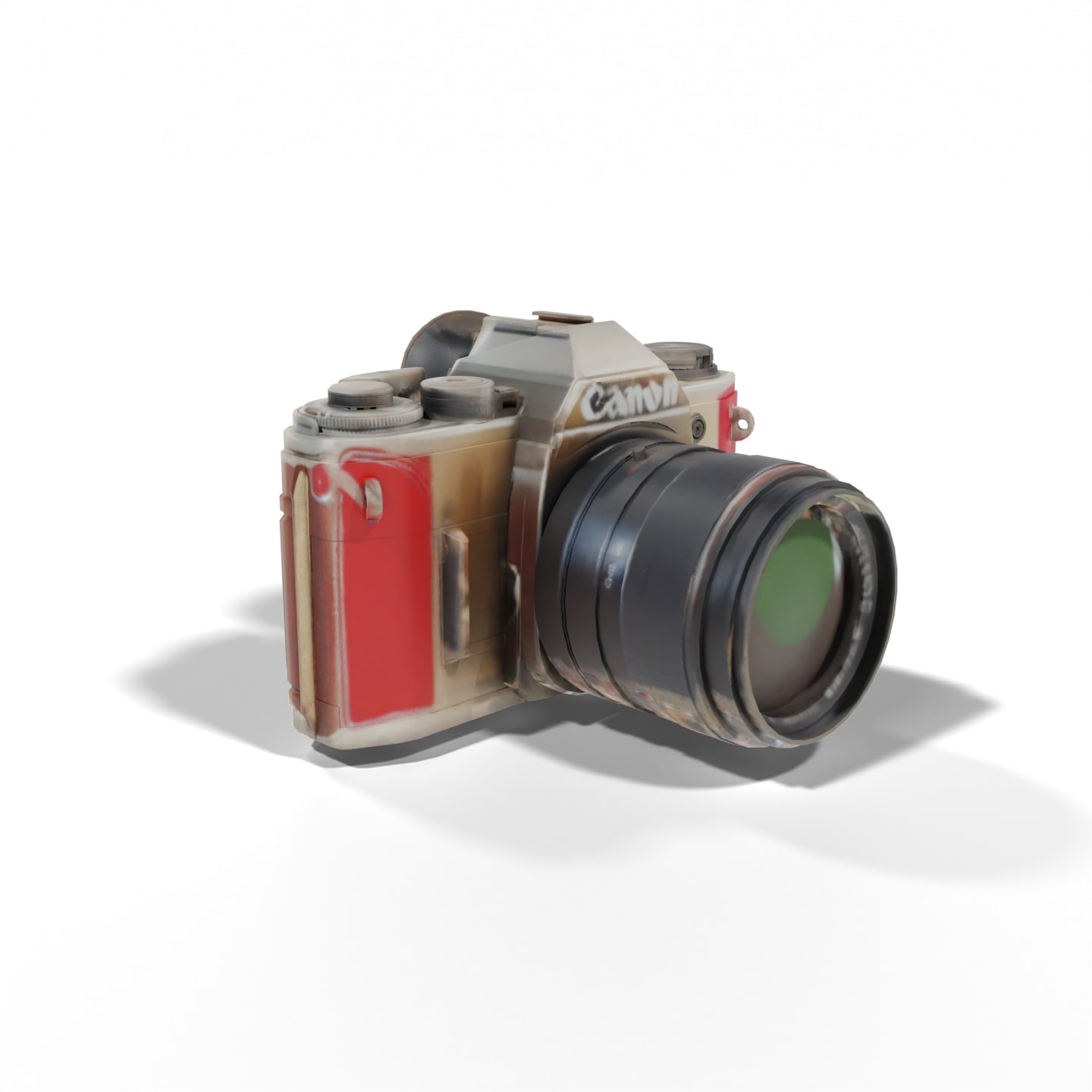}\\
\multicolumn{2}{c}{Albert Einstein, full color}&
\multicolumn{2}{c}{a Canon AT-1 Retro camera}\\
\includegraphics[width=0.2\linewidth,trim=300px 300px 300px 300px,clip]{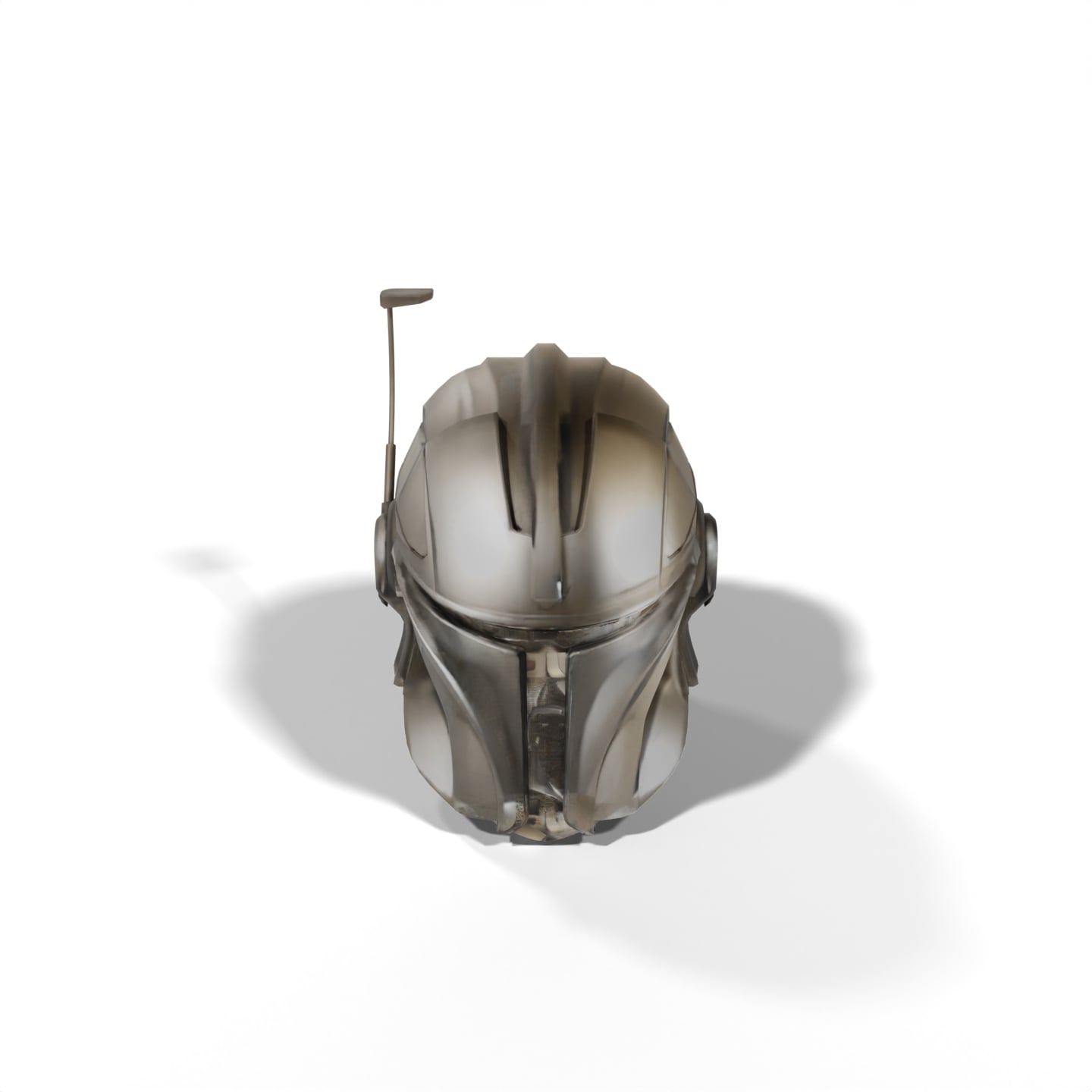}&
\includegraphics[width=0.2\linewidth,trim=300px 300px 300px 300px,clip]{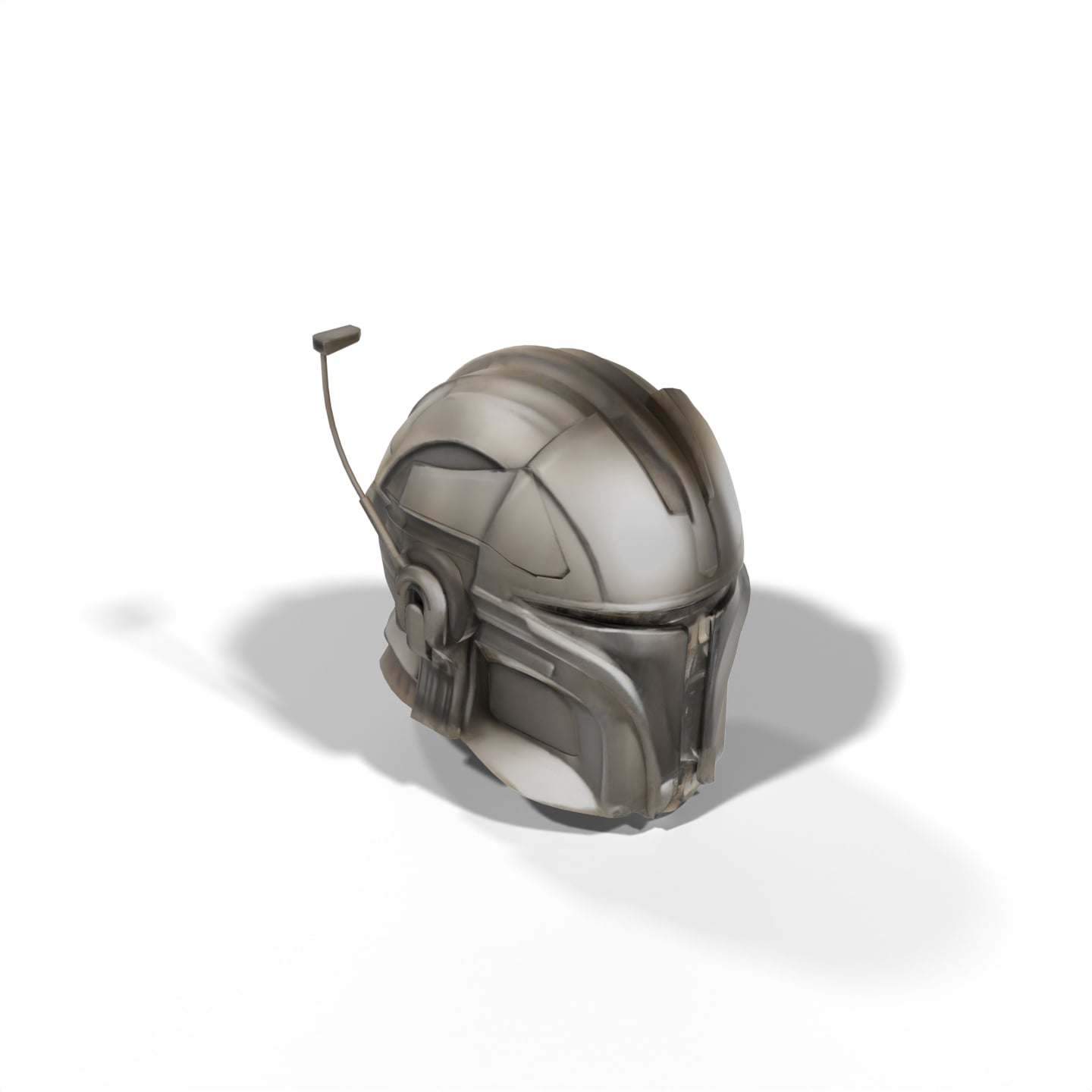}&
\includegraphics[width=0.2\linewidth,trim=300px 300px 300px 300px,clip]{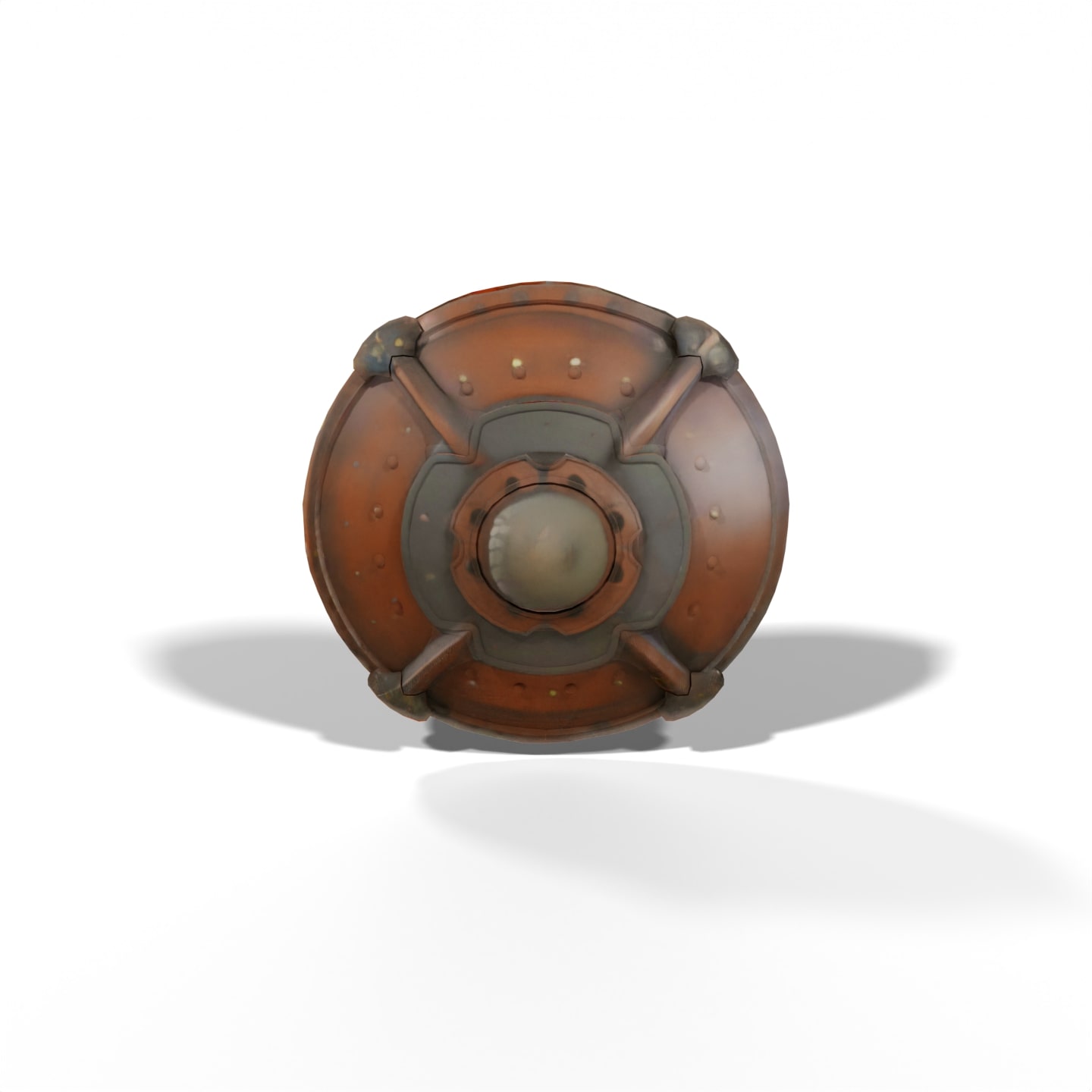}&
\includegraphics[width=0.2\linewidth,trim=300px 300px 300px 300px,clip]{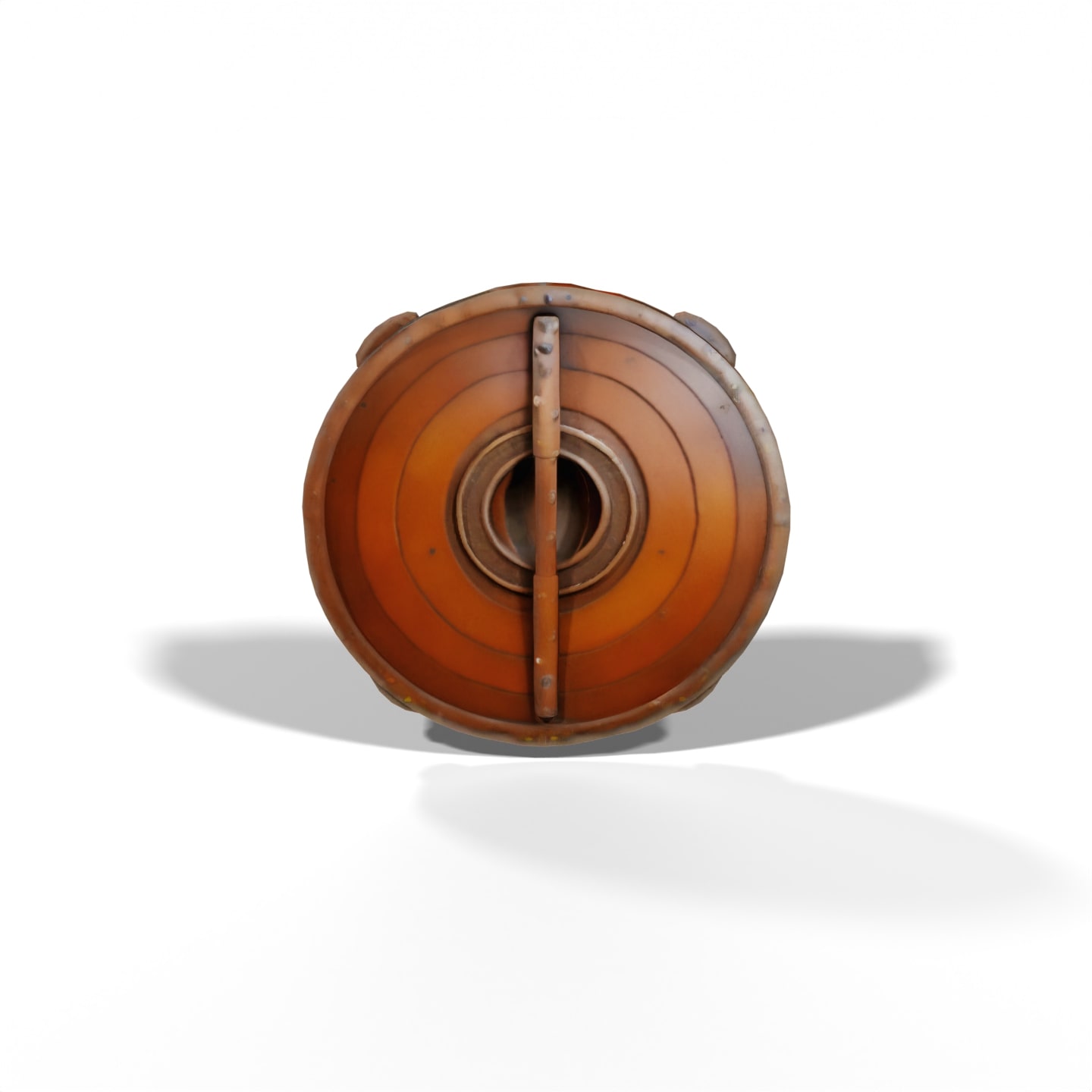}\\
\multicolumn{2}{c}{Mandalorian helmet, Star War}&
\multicolumn{2}{c}{wooden shield adorned with iron embellishments}\\
\end{tabular}
\caption{\label{fig:teaser} We have evaluated our method on a row of 3D models and our method consistently generates high-quality texture images with multi-view consistency, some of which are illustrated with two views for each model, along with the text prompt below.}
\end{teaserfigure}

\makeatletter
\let\@authorsaddresses\@empty
\makeatother

\maketitle

\section{Introduction}
Automatic content creation is one of the ultimate goals of computer graphics. Decades of efforts, including conventional content creation techniques~\cite{prusinkiewicz1996systems,muller2006procedural,chen2008interactive}, have been invested in this domain.
The most recent success of Latent Diffusion Model (LDM)~\cite{rombach2022LDM} trained on large-scale internet image dataset significantly advances the multi-model expressivity of generative models. Since then, continued efforts have been made to extend the expressivity from 2D images to 3D models. However, due to the limited dataset and computational resources, 3D diffusion models~\cite{zhou20213d,vahdat2022lion} still cannot achieve similar variety and scalability as their 2D counterpart. In parallel, researchers have turned to extracting 3D information from pre-trained 2D LDM. An exemplary technique is DreamFusion~\cite{poole2022dreamfusion}, which optimizes a neural radiance field using Score Distillation Sampling (SDS). 
This technique, however, tends to produce appearances with unnatural colors for the geometry.
To bridge this gap, the latest texture generation techniques~\cite{richardson2023TEXTure,cao2023texfusion,chen2023text2tex} propose to distill texture images for given 3D models from depth-conditioned, pre-trained 2D LDM.

Our work aims at extracting texture images of a consistently higher quality from pre-trained 2D LDM. Our key technical challenge lies in resolving the multi-view consistency, i.e., the texture should lead to semantically and visually consistent rendered appearance across all camera views. Regretfully, existing techniques~\cite{richardson2023TEXTure,cao2023texfusion,chen2023text2tex} still cannot achieve satisfactory results due to several reasons. First, the early approaches~\cite{richardson2023TEXTure,chen2023text2tex} runs the denoising process for each view, sequentially, leading to sub-optimal results. Later,~\citet{cao2023texfusion} tackle this problem by tightly coupling the denoising process with multi-view fusion. They choose to fuse the texture in latent space using noisy screen-space multi-view images. However, since latent codes for different views are noised in separate diffusion processes, such manipulation can be detrimental to the image quality and even counteract the consistency. Further, these methods introduce unnecessary assumptions between multiple views. For example, \citet{richardson2023TEXTure,chen2023text2tex} assume the next camera view can either keep, refine, or overwrite the texture generated from the previous camera view. While~\cite{cao2023texfusion} assumes the rendered images from multiple views have sequential dependence. This assumption is counter-intuitive, because difference camera views are not directly correlated through a common texture image. 

We propose a new approach for generating a multi-view consistent texture image from a given 3D model and a text prompt. As the key point of departure from prior works, our method avoids direct operations on the latent codes and does not use assumptions on inter-view correlation. Instead, we fuse multiple views by joint optimization of the latents. Our technique is based on the mechanism of the celebrated DDIM scheme \cite{song2022DDIM}. During each denoising step, DDIM estimates the ultimate noiseless latent state to predict the next noise level. We then decode these noiseless latents into color space and blend them to form a texture image. We choose to update the DDIM latents across all views by optimization, such that they generate the same image as one rendered using the blended texture image. As such, we also eliminate the aforementioned assumption on sequential dependence. Extended experiments and user study confirms that our simple approach achieves consistent improvements on texture quality. 

\section{Related Work}
We review related work on the automated generation of 3D contents, including both geometry and appearance.

\subsection{Classical 2D/3D Content Generation}
In the early stage, content generation techniques are generally confined to a specific category of geometry and appearance models. For example, texture synthesize methods~\cite{wei2009state} extend a small exemplary texture tile to larger textures. Texture transfer methods~\cite{mertens2006texture} replicate the texture across different 3D models.
~\cite{chen2022auvnet} introduces a network designed to transfer texture from an example.
Procedural methods can generate both 2D textures~\cite{dong2020survey} and 3D geometries~\cite{smelik2014survey}. 
For certain categories of models, such as trees~\cite{sun2009intelligent} and buildings~\cite{talton2011metropolis, Bao2013PFV}, or branch structures~\cite{sibbing2010ISB}, specialized techniques can be designed to generate both geometry and appearance with significant varieties. 
However, all these techniques potentially suffer from two common drawbacks. First, these methods can only generate contents varied in low-level details, while the high level semantics must be kept fixed. Second, they require considerable domain knowledge to use, leading to a non-trivial learning curve.

\subsection{Learnable 3D Generative Models}
Deep learning techniques have been applied to content generation and achieved significant success in the past decade. Their success is backed by the everlasting efforts to search for powerful generative models that can efficiently represent complex multi-model distributions, of which two representative models are Variational AutoEncoders (VAE) and Generative Adversarial Networks (GAN). While originally experimented on 2D image datasets, they have been extended in~\cite{brock2016generative,wu2016learning} to represent 3D geometry via voxel grid representation. However, these works are focused on generating only 3D geometry, instead of full appearance models. This gap is bridged by several follow-up works that synthesize appearance models. For example, Text2Mesh~\cite{michel2021text2mesh} infers stylish mesh textures and displacements to match a given text prompt. 
Texture fields~\cite{oechsle2019texture} and Texturify~\cite{siddiqui2022texturify} generate the texture for a given 3D mesh to match the appearance of a 2D image. They use VAE to encode the input cues and GAN-loss to optimize the texture.
~\citet{yu2021LTG} and Mesh2Tex~\cite{bokhovkin2023mesh2tex} achieve decorating of existing 3D shapes by training a conditional texture generator, leveraging GAN.
While the aforementioned techniques are focused on generating the color field, TANGO~\cite{chen2022tango} goes beyond this paradigm to generate a complete BRDF model for an existing 3D object, where the generated BRDF model is matched with a text prompt using the CLIP loss. Unlike all these works that generate appearance for given geometry, GET3D~\cite{NEURIPS2022GET3D} jointly generate textured meshes. They parametrically represent the geometry and color through auto-encoding, and use GAN to train the joint generative model. 
Similarly, ShaDDR~\cite{Chen2023shaddr} construct both the geometry and texture from a provided example drawing inspiration from GAN.
Compared with the more recent LDM, VAE and GAN has limited expressivity. As a result, separate models oftentimes need to be trained for each category of data, which limits their domain of usage.

\subsection{2D/3D Diffusion Models}
Diffusion models~\cite{ho2020DDPM, song2021scorebased} demonstrate significantly improved ability to approximate complex distributions. Based on this model, there are many advances in the area of 2D image generation over the past two years. In particular, the celebrated LDM~\cite{rombach2022LDM} produces high-quality images with affordable memory and computation. In addition, the multi-model conditioning of LDM, using text and depth images, significantly improves their amenability to non-expert users. A key technique behind the conditional generation is the blending between classified and classifer-free guidance~\cite{ho2022CFG}. In parallel, a series of methods are designed to accelerate the sampling of the reverse diffusion process~\cite{song2022DDIM,lu2022DPM}.

Extending from 2D to 3D content generation, there are several attempts that train diffusion models to directly generate 3D assets~\cite{müller2023DiffRF, karnewar2023HoloDiffusion,zeng2022lion}. For appearance generation, in particular, Point-UV~\cite{yu2023Point-UV} is a 3D diffusion model that predicts the color of a given mesh and synthesises the texture. However, the variety and quality of these generated 3D contents are considerably lower than their 2D counterparts~\cite{rombach2022LDM}. This is largely because 3D content generation requires models with considerably more parameters, while the amount of computational resources and available datasets are severely inadequate. 

\vspace{-2mm}
\subsection{3D Content Distillation from 2D LDM}
Considering the inherent difficulty to train full-fledged large 3D LDM, researchers have considered extracting 3D contents from pre-trained 2D diffusion model~\cite{von-platen-etal-2022-diffusers}. 
The seminal work of DreamFusion~\cite{poole2022dreamfusion} proposes SDS to optimize a neural radiance field~\cite{mildenhall2020nerf} from pre-trained models. Through the optimization, DreamFusion generates colored geometry conforming to the input text prompt. There are also other SDS-based methods with further improved generation quality on geometry and/or appearance~\cite{Chen_2023_Fantasia3D,lin2023magic3d,metzer2022latent-nerf}. For higher quality or better alignment with input condition, several methods even fine-tuned the pre-trained LDM~\cite{wang2023prolificdreamer,yu2023boosting3d}. However, an noticeable limitation of SDS is the excessive usage of classifier-free guidance weighting~\cite{ho2022CFG}, which suffers from a low fidelity with over-saturation and overexposure.
Besides, DG3D~\cite{zuo2023dg3d} also incorporates a diffusion model to produce textured meshes, although the appearance quality remains constrained.

Instead of generating both geometry and appearance all at once, the depth-conditioned LDM enables the application of generating appearance for a given geometry. To conquer this seemingly easier task, the major technical challenge lies in multi-view consistency, i.e., the appearance must be consistent across all possible camera views. To enforce such consistency, an intuitive idea is to generate images from a set of sampled views and then ``paint'' them onto the 3D model. \citet{richardson2023TEXTure} and \citet{chen2023text2tex} propose their texture generation methods based on this idea, where each new camera view revises the overlapping parts of the texture image from the previous view. However, this method still suffers from various artifacts including seam, noise, and meaningless fragments.
Several texture painting works enhance consistency within their respective contexts. Paint3D~\cite{zeng2023paint3d} trains a dedicated model to fill incomplete areas during the texture painting. MVDiffusion~\cite{Tang2023mvdiffusion} focuses on panorama generation with a given mesh and fine-tune a 2D diffusion model to maintain consistency.
To alleviate the inconsistency more directly, \citet{cao2023texfusion} propose to sequentially correlate the noised images from all different views. This method operates entirely in the latent space and finally use an additional neural field optimization to reconstruct the color-space texture image. In our experiments, however, their latent-space manipulations can reduce the quality of the texture image. By comparison, we adopt the multi-view fusion in the color space and then solve an optimization during each denoising step to update the latents, which further eliminates the assumption on sequential inter-view correlation.

\section{Preliminaries}

\subsection{\label{sec:problem}The Texture Painting Problem}
Our intended problem takes the same form as prior works~\cite{richardson2023TEXTure,cao2023texfusion,chen2023text2tex}. We are given a 3D object represented by a mesh $\mathcal{M}=\langle\mathcal{V},\mathcal{F}\rangle$, with $\mathcal{V}$ and $\mathcal{F}$ being a set of vertices and facets, respectively. 
We further assume the appearance of the mesh is determined by a texture image and a UV map. The UV map defines an almost everywhere injective function $\mathcal{T}: \mathbb{R}^2\mapsto\mathbb{R}^3$ mapping a point on the 2D plane to that of the 3D surface on $\mathcal{M}$, which assigns the texture color $I(u)$ to the surface point $\mathcal{T}(u)$. In addition to the surface mesh $\mathcal{M}$, we further assume users provide a text-based description of the mesh semantics and appearance, which is converted to a text prompt embedding $h$ as the conditional input.
Given $\mathcal{M}$ and $h$, our goal is to automatically infer the texture image $I$ that has consistent appearance and semantic meaning under arbitrary camera views. 

\subsection{\label{sec:diffusion}Diffusion Model and Denoising Procedures}
Our method is based on the pre-trained, depth-conditioned LDM~\cite{rombach2022LDM} for generating 2D images, guided by a text prompt. The diffusion model~\cite{sohl2015deep} is a generative model inspired by thermal dynamics. First, we define the diffusion process that gradually injects Gaussian noise into the data distribution $z_0\sim p(z_0)$, leading to the following forward Markovian model
$p(z_{1:T}|z_0)=\prod_{t=0}^Tp(z_{t+1}|z_t)$
with $p(z_{t+1}|z_t)=\mathcal{N}\left(\sqrt{\frac{\alpha_t}{\alpha_{t-1}}}z_t,\left(1-\frac{\alpha_t}{\alpha_{t-1}}\right)I\right)$ and $\alpha_{i}\in(0,1]$ 
being a decreasing schedule of noise coefficients. In the forward process, the data distribution is blended into a pure Gaussian noise at the $t$th timestep. The diffusion model works by learning the inverse of the diffusion process that gradually removes noise from $z_t$. Taking the original Denoising Diffusion Probabilistic Models (DDPM)~\cite{ho2020DDPM} for example, the noise component in $z_t$ is predicted using a neural network $\epsilon_\theta(z_t,t,h)$, and we can sample the next level of less noisy version $z_{t-1}$ by:
\begin{align}
\label{eq:DDPM}
z_{t-1}\sim\mathcal{N}\left(\frac{1}{\sqrt{\alpha_t}}\left(z_t-\frac{\beta_t}{\sqrt{1-\bar{\alpha}_t}}\epsilon_\theta(z_t,t,h)\right),\sigma_tI\right),
\end{align}
with $\beta_t=1-\alpha_t$, $\bar{\alpha}_t=\prod_{s=1}^t\alpha_s$, and $\sigma_t$ being the noise variance at level $t$. Here $h$ is some latent code encoding the user-input guidance information. The original DDPM has been improved in its expressivity and inference efficacy in various prior works, of which the most outstanding results is LDM~\cite{rombach2022LDM} that propose to perform the inverse process in the latent space.
We use their notation and denote the latent code as $z$ and the original image as $x$. The pre-trained autoencoder is denoted as: $z=\mathcal{E}(x)$ and $x=\mathcal{D}(z)$. In parallel, DDPM suffers from slow inference due to its temporally sequential nature of the forward process. In view of this, the more recent DDIM~\cite{song2022DDIM} proposes to improve the inference cost by using a non-Markovian inverse process. 
The denoising step of DDIM in Equation~\ref{eq:DDPM} is replaced with:
\begin{align}
\label{eq:DDIM}
z_{t-1}&\sim\mathcal{N}\left(\sqrt{\alpha_{t-1}}\hat{z}_{0,t}+\sqrt{1-\alpha_{t-1}-\sigma_t^2}\epsilon_\theta(z_t,t,h),\sigma_tI\right)\nonumber\\
\hat{z}_{0,t}&\triangleq\frac{z_t-\sqrt{1-\alpha_t}\epsilon_\theta(z_t,t,h)}{\sqrt{\alpha_t}},
\end{align}
where we use $\hat{z}_{0,t}$ to denote the predicted noiseless data distribution at $t$th timestep. We will use this property to design our algorithm to enforce multi-view consistency.

\subsection{Multi-view Inconsistency in Texture Painting}
We detail the texture generation process to infer $I$ from $\mathcal{M}$ and $h$. During painting, we query the 2D LDM using text prompt $h$ from a set $\mathcal{C}$ of sampled camera view positions denoted as $\mathcal{C}=\{c^i\}$, where we use superscript to denote camera indices throughout. We denote $\mathcal{R}(\mathcal{M},I,c)$ as a rendering function, and $\mathcal{R}_z, \mathcal{R}_x$ as renderers working in the latent and color space, respectively. Similarly, we denote $I_z$ and $I_x$ as the texture image in the corresponding space.

\begin{figure}[t]
\centering
\setlength{\tabcolsep}{0px}
\begin{tabular}{ccc}
&
\includegraphics[width=.32\linewidth]{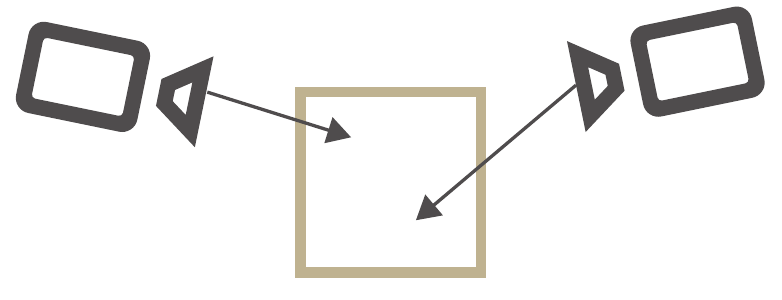}
\put(-75,8){$c^1$}
\put(-10,10){$c^2$}
\put(-50,25){texels}&
\includegraphics[width=.32\linewidth]{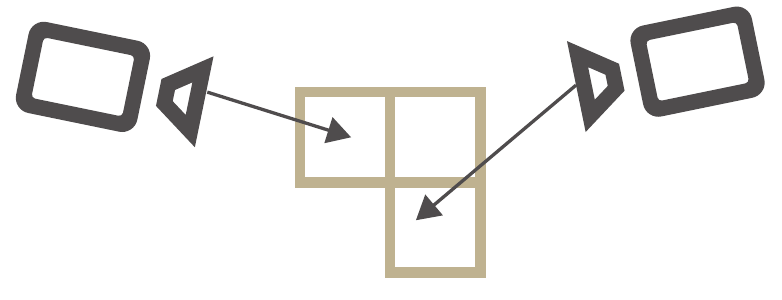}	
\put(-75,8){$c^1$}
\put(-10,10){$c^2$}
\put(-50,25){texels}\\
\includegraphics[width=.32\linewidth,
trim=0 150px 0 160px,clip]{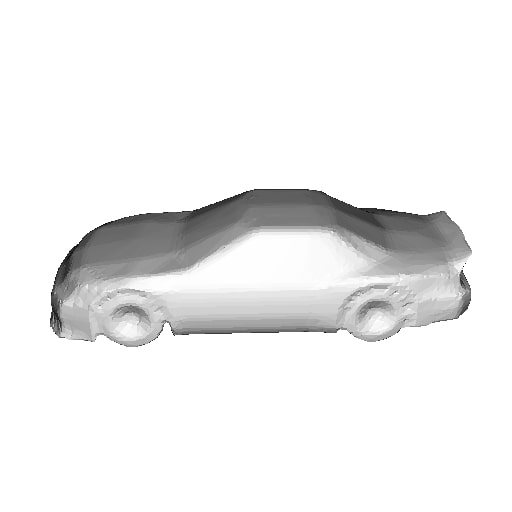}&
\includegraphics[width=.32\linewidth,
trim=0 150px 0 160px,clip]{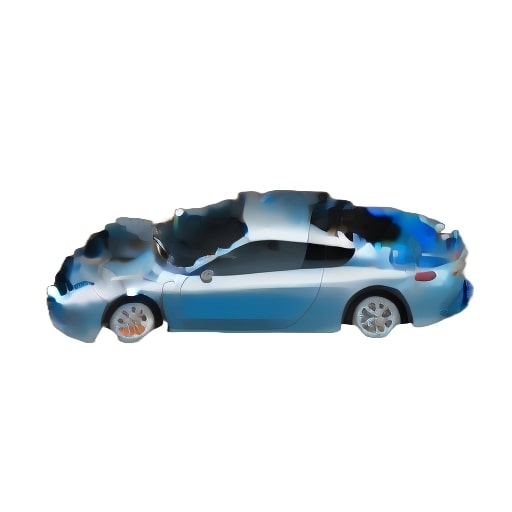}&
\includegraphics[width=.32\linewidth,
trim=0 150px 0 160px,clip]{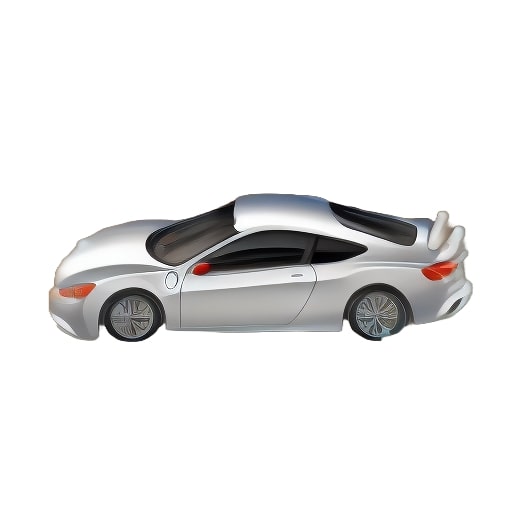}\\
view 1 &
low-res $I_{z,0}\to z_0^1$ &
high-res $I_{z,0}\to z_0^1$\\
\includegraphics[width=.32\linewidth,
trim=0 70px 0 160px,clip]{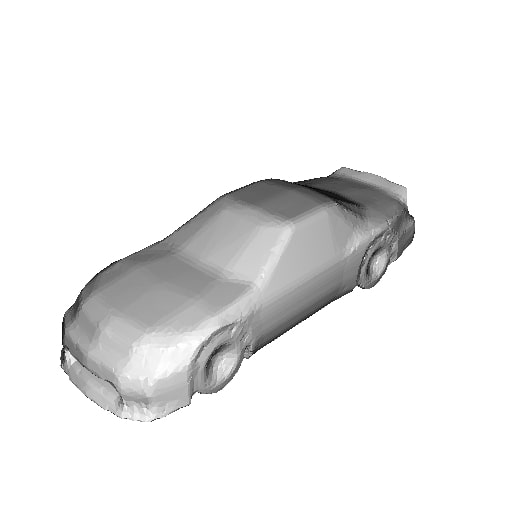}&
\includegraphics[width=.32\linewidth,
trim=0 70px 0 160px,clip]{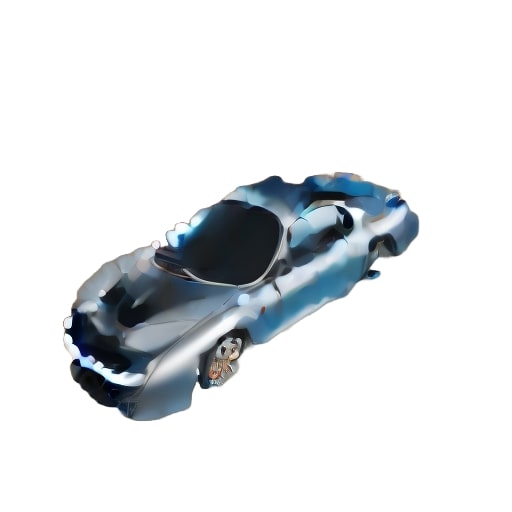}&
\includegraphics[width=.32\linewidth,
trim=0 70px 0 160px,clip]{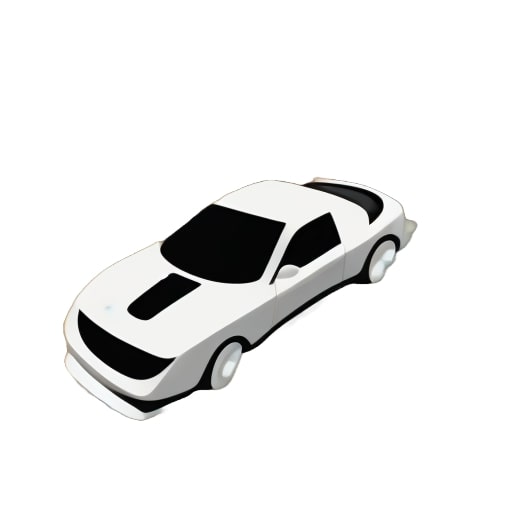}\\
view 2 &
low-res $I_{z,0}\to z_0^2$ &
high-res $I_{z,0}\to z_0^2$\\
\end{tabular}

\caption{\label{fig:pitfall} We run diffusion processes from two nearby views and enforce consistency by blending the noisy latent code into $I_z$ during each step. Under a low-res $I_z$, two views are correlated by sampling largely the same set of texels, thus achieving multi-view consistency, but low-res $I_z$ leads to low-quality blurry images (middle). Instead, clear images are derived under a high-res $I_z$, but the two views can fetch entirely different sets of texels due to the nearest sampling scheme used by SIMS, failing to achieve consistency (right).}

\vspace{-3mm}
\end{figure}

To solve the texture painting problem, prior methods run a separate denoising process for each $c^i$ using LDM, generating a sequence of latent images $z_t^i$. We denote the updated latent texture using the first $l$ camera views as $I_z^l$, then all these prior works assume a sequential dependence on camera views, i.e., the following probabilistic model:
\begin{align}
\label{eq:sequential}
p(I_{z,x}^j)=p(I_{z,x}^0)\prod_{k=1}^{j}p(I_{z,x}^k|I_{z,x}^{k-1}),
\end{align}
either in latent or color space, as denoted by subscript. The first category of methods~\cite{chen2023text2tex,cao2023texfusion} run the denoising process for each view and update the color texture sequentially, i.e., assuming subscript $x$ in Equation~\ref{eq:sequential}. 
Later,~\citet{cao2023texfusion} noticed that sequential denoising leads to various artifacts. Instead, they proposed to couple the denoising process with texture fusion by sequentially merging noisy images into a latent texture during each step, i.e., assuming subscript $z$ in Equation ~\ref{eq:sequential}. 
To mitigate the inconsistency across multi-views in each denoising step, they use a meticulously designed update rule named SIMS.
Unfortunately, as illustrated in Figure~\ref{fig:pitfall}, we show that this method can still result in blurred images under a low-resolution $I_z$ or inconsistent images using a high-resolution $I_z$, leading to a dilemma in choosing appropriate texture resolution.

\section{Method}
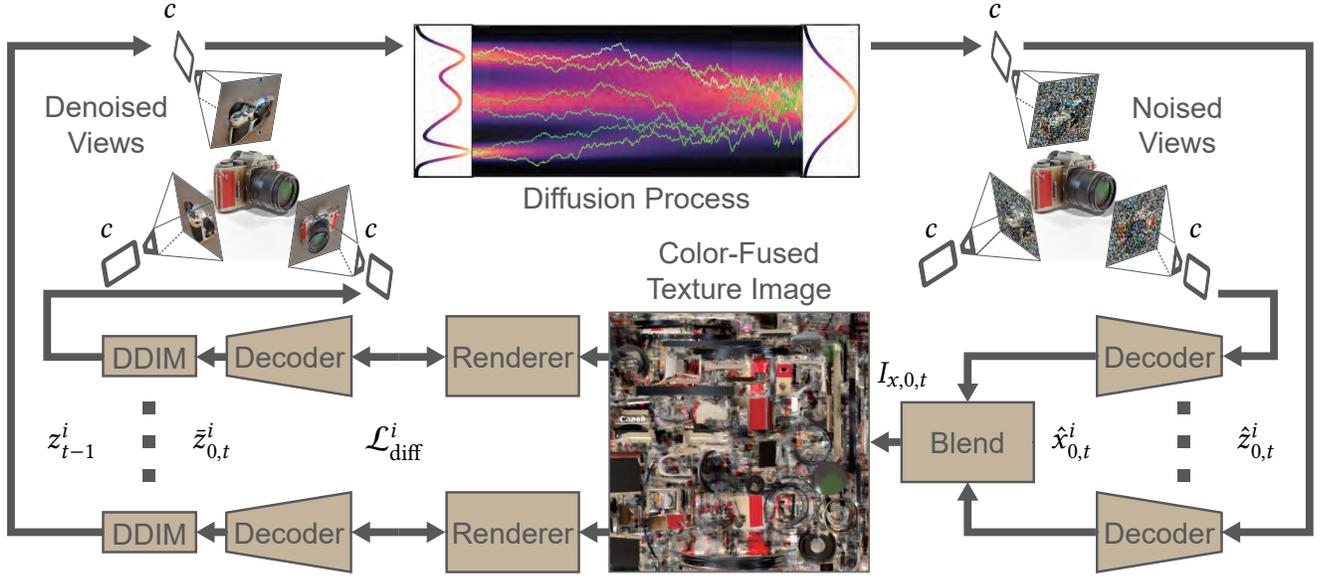
\begin{figure*}[t]
\centering
\includegraphics[width=\linewidth]{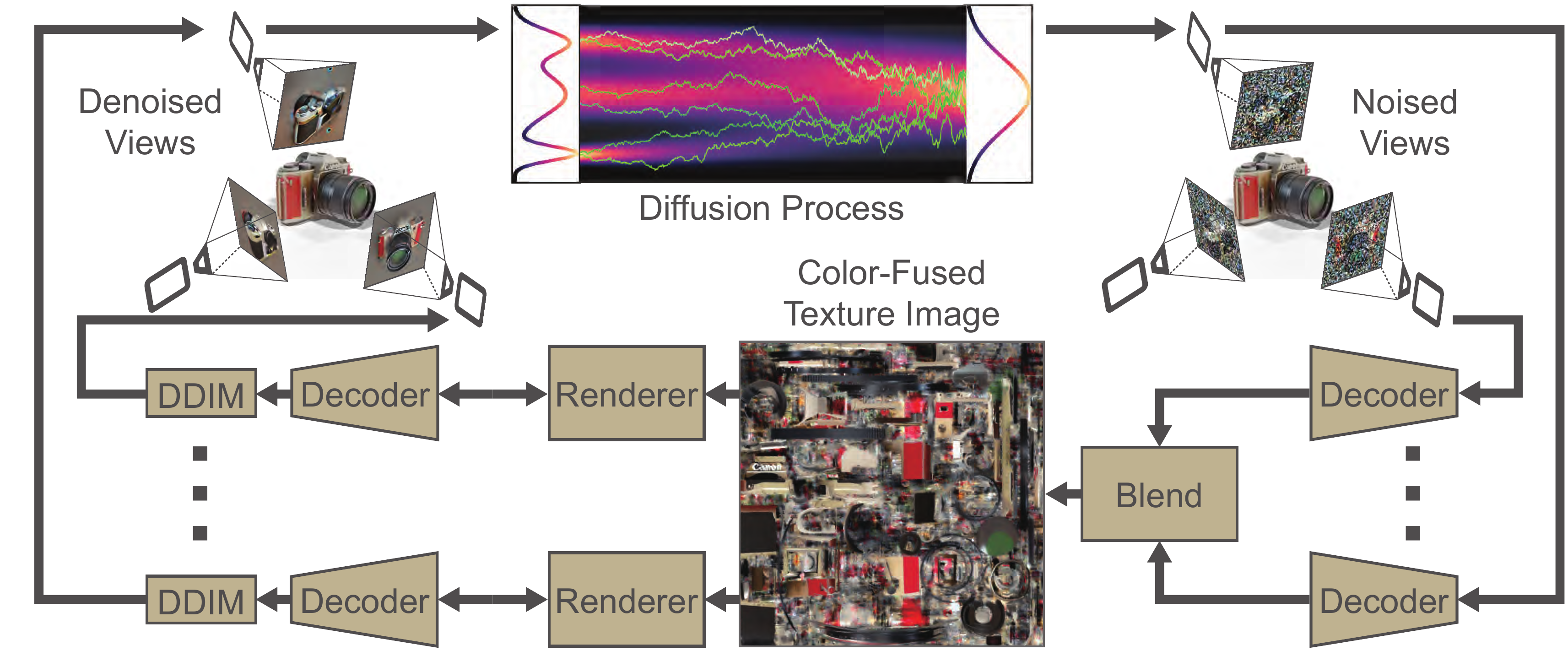}
\put(-30,50){\LARGE{$\hat{z}_{0,t}^i$}}
\put(-100,50){\LARGE{$\hat{x}_{0,t}^i$}}
\put(-360,50){\LARGE{$\mathcal{L}_\text{diff}^i$}}
\put(-425,50){\LARGE{$\bar{z}_{0,t}^i$}}
\put(-480,50){\LARGE{$z_{t-1}^i$}}
\put(-166,75){\LARGE{$I_{x, 0, t}$}}
\put(-124,213){\LARGE{$c$}}
\put(-436,213){\LARGE{$c$}}
\put(-148,130){\LARGE{$c$}}
\put(-460,130){\LARGE{$c$}}
\put(-48,130){\LARGE{$c$}}
\put(-360,130){\LARGE{$c$}}
\caption{\label{fig:pipeline}Our modified multi-DDIM procedure that enforces multi-view consistency. Each view runs a separate denoising procedure using DDIM scheme. For each denoising step, DDIM predicts a latent code $\hat{z}_{0,t}^i$ for the $i$th view at $0$th timestep. These $\hat{z}_{0,t}^i$ are decoded to the color space, yielding $\hat{x}_{0,t}^i$. We then blend these views into a common color-space texture image by weighted averaging. Next, we perform an optimization to update $\hat{z}_{0,t}^i$ into $\bar{z}_{0,t}^i$ for all views, such that their decoded images match their corresponding rendered views using the blended texture image. These updated latent codes are then plugged into DDIM to predict the next noise level.}
\end{figure*}

Our texture generator pipeline is illustrated in Figure~\ref{fig:pipeline}, which is a modified multi-DDIM procedure~\cite{song2022DDIM} that enforces multi-view consistency. In this section, we detail our modified DDIM procedure in Section~\ref{sec:DDIM}. 
Then, we propose two extensions to our pipeline in Section~\ref{sec:extensions}.

\subsection{\label{sec:DDIM}DDIM with Multi-view Consistency}
To circumvent the pitfall of prior texture painting methods, we make two observations. First, although $z_t^i$ during an intermediary denoising step has drastically different noise component across views, the $\hat{z}_{0,t}^i$ predicted by the DDIM scheme is an estimation at the $0$th timestep, which is supposed to be noiseless. Second, we notice that the auto-encoder used by LDM~\cite{rombach2022LDM} can be highly nonlinear map, and consistency in the color space does not imply that in the latent space. Therefore, we propose to enforce multi-view consistency by modifying $\hat{z}_{0,t}^i$ in the color space. As illustrated in Figure~\ref{fig:pipeline}, specifically, we perform denoising process for all the sampled views in parallel. For the $t$th timestep, our method yields the set of latent images $\{z_t^i\}$. We then predict $0$th timestep using Equation~\ref{eq:DDIM} to yield $\{\hat{z}_{0,t}^i\}$. 

Next, we switch from latent to color space by applying the decoder, yielding predicted $0$th timestep color images:
$\hat{x}_{0,t}^i=\mathcal{D}(\hat{z}_{0,t}^i)$.
In the color space, we then fuse the images to a color texture $I_{x,0,t}$. To this end, we use simple weighted averaging scheme. Specifically each pixel $u$ lies on some facet with its outer-normal denoted as $n(u)$. The camera direction from $i$th view to $\mathcal{T}(u)$ is denoted as $c^i-\mathcal{T}(u)$. We assign the following weight to the $i$th view:
\begin{align*}
w^i=\max\left(0,\left\langle\frac{c^i-\mathcal{T}(u)}{\|c^i-\mathcal{T}(u)\|},n(u)\right\rangle\right),
\end{align*}
which assigns larger weights to views that are nearly orthogonal to the facet and cover more pixels. We then assign the color texture via the following scheme:
\begin{align}
\label{eq:FuseX}
I_{x,0,t}(u)=\sum_{i\in\mathcal{C}(\mathcal{T}(u))}w^i \hat{x}_{0,t}^i(\mathcal{T}(u))/
\sum_{i=1}^{|\mathcal{C}|}w^i,
\end{align}
where we use $\mathcal{C}(\mathcal{T}(u))$ to denote all the cameras that are visible from $\mathcal{T}(u)$. Here we slightly abuse notation to use $\hat{x}_{0,t}^i(\mathcal{T}(u))$ to denote the sampling operator that fetches the pixel corresponding to world space coordinate $\mathcal{T}(u)$. Following~\cite{cao2023texfusion}, we use nearest neighbor scheme to perform the pixel interpolation.

\begin{algorithm}[t]
\caption{TexPainter}
\label{alg:TexPainter}
\begin{algorithmic}
\STATE\textbf{Input:} Mesh $\mathcal{M}$, text prompt $h$
\STATE\textbf{Output:} Color texture image $I$
\STATE Sample $\{z_T^i\sim\mathcal{N}(0,I)\}$
\FOR{$t=T,\cdots,0$}
\STATE Predict $\{\epsilon_\theta(z_t^i,t,h)\}$ using depth-conditioned LDM
\STATE Predict $\{\hat{z}_{0,t}^i\}$ (Equation~\ref{eq:DDIM})
\STATE \textcolor{brown}{Decode $\{\hat{x}_{0,t}^i=\mathcal{D}(\hat{z}_{0,t}^i)\}$}
\STATE \textcolor{brown}{Color-space fuse $I_{x,0,t}$} (Equation~\ref{eq:FuseX})
\IF{$t=0$}
\STATE Return $I_{x,0,0}$
\ENDIF
\STATE \textcolor{brown}{Optimize $\{\bar{z}_{0,t}^i\}$} (Equation~\ref{eq:FuseZ})
\STATE Apply DDIM to yield $\{z_{t-1}^i\}$ (Equation~\ref{eq:DDIM_multi_view})
\ENDFOR
\end{algorithmic}
\end{algorithm}

With the color-fused texture image $I_{x,0,t}$, we now switch back from color to latent space, updating $\{\hat{z}_{0,t}^i\}$ to yield a set of adjusted latent codes $\{\bar{z}_{0,t}^i\}$ with multi-view consistency. To this end, we use an optimization to enforce that the updated $\{\bar{z}_{0,t}^i\}$ yield consistent images under corresponding views. Our optimization takes the following form:
\begin{equation}
\begin{aligned}
\label{eq:FuseZ}
&\underset{\{\bar{z}_{0,t}^i\}}{\text{argmin}}\sum_{i=1}^{|\mathcal{C}|}\mathcal{L}_\text{diff}(\bar{z}_{0,t}^i,c^i)\\
&\mathcal{L}_\text{diff}(z,c)\triangleq\|\mathcal{D}(z)-\mathcal{R}(\mathcal{M},I_{x,0,t},c)\|_1,
\end{aligned}
\end{equation}
where we use $\{\hat{z}_{0,t}^i\}$ as our initial guess. Note that the rendering function in this optimization can be pre-evaluated before optimization. Therefore, the optimization does not require a differentiable renderer as used in~\cite{cao2023texfusion} and only involves the derivatives of $\mathcal{D}$, which is relatively efficient to compute. Note that Equation~\ref{eq:FuseZ} implies all camera views are treated equally without sequential dependency. Finally, we plug the updated $\{\bar{z}_{0,t}^i\}$ into DDIM, obtaining:
\begin{align}
\label{eq:DDIM_multi_view}
z_{t-1}^i&\sim\mathcal{N}\left(\sqrt{\alpha_{t-1}}\bar{z}_{0,t}^i+\sqrt{1-\alpha_{t-1}-\sigma_t^2}\epsilon_\theta(z_t^i,t,h),\sigma_tI\right).
\end{align}
We summarize our method in Algorithm~\ref{alg:TexPainter}, where our color-fusion procedure is highlighted in brown. As a computational drawback, our method requires solving the optimization Equation~\ref{eq:FuseZ} per denoising step, while prior work~\cite{cao2023texfusion} only applies the optimization once to reconstruct the color texture at last. Fortunately, since our objective function only requires back-propagation through the decoder $\mathcal{D}$, which is quite efficient, the cost of optimization is still manageable.

\subsection{Extensions}
\label{sec:extensions}
\begin{figure}[t]
\centering
\includegraphics[width=0.8\linewidth,trim=0 50px 0px 0px,clip]{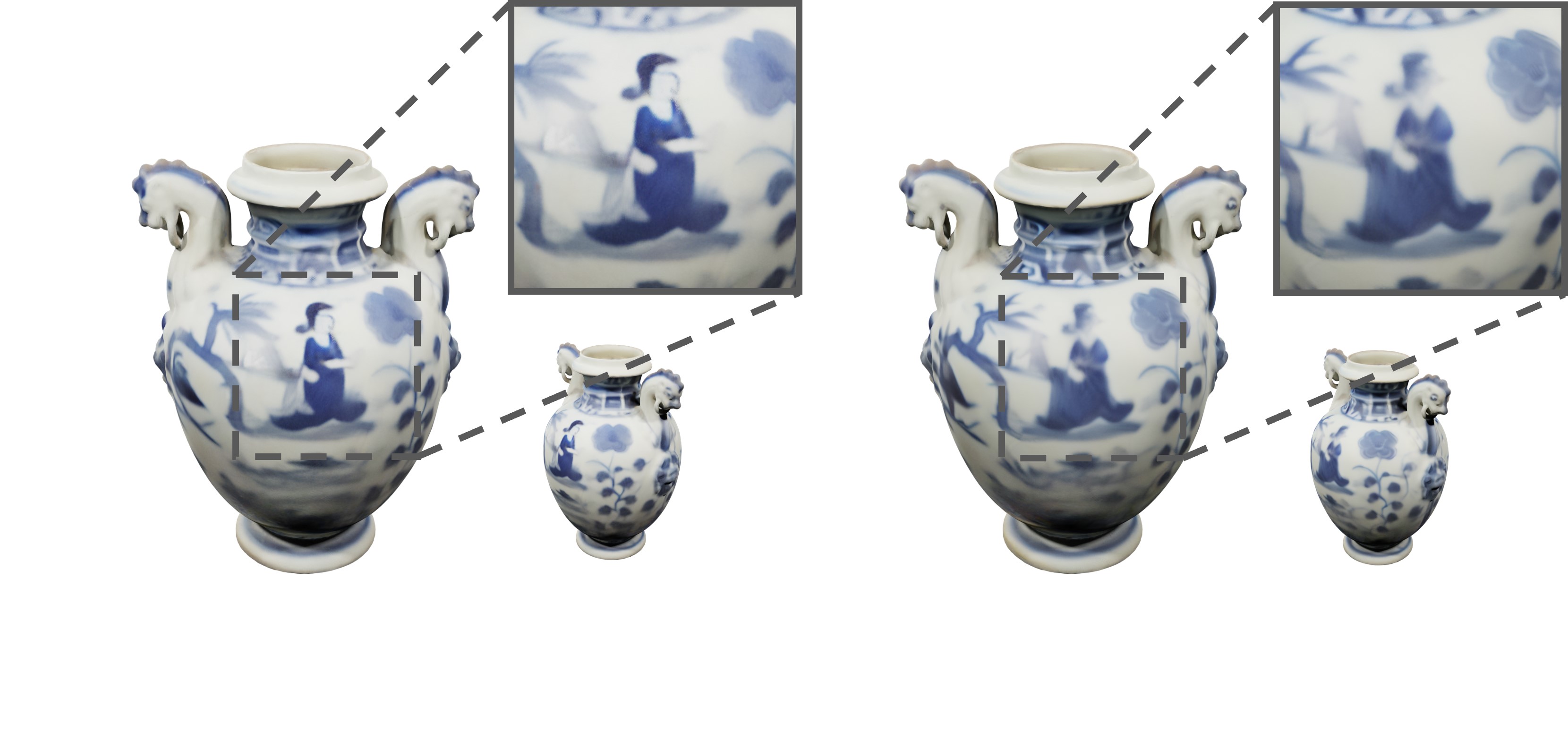}
\caption{\label{fig:jointOpt}We highlight the benefits of our joint optimization Equation~\ref{eq:FuseZFineTune} (left) as compared with Equation~\ref{eq:FuseX} (right). The joint optimization achieves better texture quality in areas not well-sampled by camera views, but increases the inference cost from 25min to 66min.}
\vspace{-3mm}
\end{figure}

We discuss two extensions to our method. First, we notice that using simple weighted averaging as in Equation~\ref{eq:FuseX} for color-fusion is a mere compromise between inference speed and texture quality. Our ultimate goal for this step is to adjust the predicted, noiseless latent codes such that they can be generated by a single unified texture image. We could achieve this by a joint optimization of $I_{x,0,t}$ and $\{\bar{z}_{0,t}^i\}$ via the following formulation:
\begin{align}
\label{eq:FuseZFineTune}
\underset{\{\bar{z}_{0,t}^i\},I_{x,0,t}}{\text{argmin}}\sum_{i=1}^{|\mathcal{C}|}\mathcal{L}_\text{diff}(\bar{z}_{0,t}^i,c^i)+\mathcal{L}_\text{diff}(\hat{z}_{0,t}^i,c^i).
\end{align}
In practice, we can still use $\{\hat{z}_{0,t}^i\}$ and Equation~\ref{eq:FuseX} as our initial guess for $\{\bar{z}_{0,t}^i\}$ and $I_{x,0,t}$, respectively, and then optimize Equation~\ref{eq:FuseZFineTune} to fine-tune. In practice, we find this procedure slightly improve the quality of the texture, especially in the areas that are not well-covered by the camera views as illustrated in Figure~\ref{fig:jointOpt}. However, since solving Equation~\ref{eq:FuseZFineTune} requires a differentiable renderer such as~\cite{laine2020modular}, involves more decision variables, and needs to be performed during every denoising step, this procedure can considerably increase the inference cost. In practice, we suggest using Algorithm~\ref{alg:TexPainter} for fast preview and use Equation~\ref{eq:FuseZFineTune} to generate final results offline.

\begin{figure}[t]
\centering
\includegraphics[width=0.8\linewidth,trim=0 400px 230px 350px,clip]{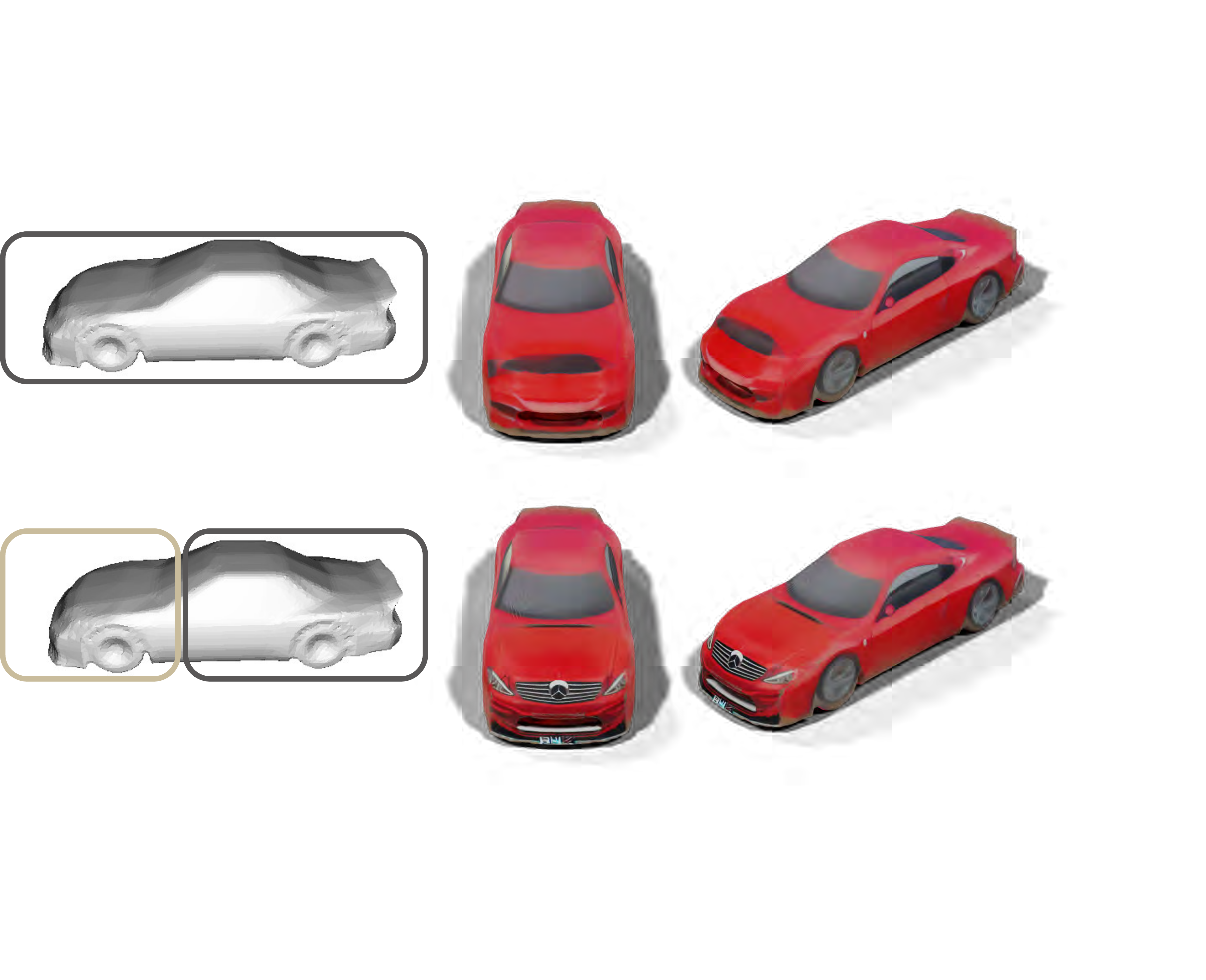}
\definecolor{myBrown}{RGB}{201,188,156}
\put(-200,60){beautiful red sports car}
\put(-200,6){\textcolor{myBrown}{red car, Benz emblem}}
\caption{\label{fig:blendTexture} A car model with its generated texture from a global prompt (top) and different prompts for different local regions (bottom).}
\vspace{-1mm}
\end{figure}

In addition, we realize that our method can be generalized into a framework to add constraints between multiple denoising process, which is not limited to multi-view consistency constraints. As an example, we show that our method can be used to achieve a similar effect as blended latent diffusion~\cite{avrahami2023blended}, but extended to 3D texture domain. To this end, we can assign two sets of cameras, but adopts two separate text prompts for each set, respectively. We then either use Equation~\ref{eq:FuseZ} to blend the texture or solve a joint optimization for two parts together using Equation~\ref{eq:FuseZFineTune}. The results are illustrated in Figure~\ref{fig:blendTexture}, where the views from two types of cameras are blended seamlessly. We believe this approach can be further extended to multiple sets of prompts or other constraints between multiple diffusion process, which is left as future work.
\begin{figure}[ht]
\centering
\setlength{\tabcolsep}{0px}
\begin{tabular}{cc}
\includegraphics[width=.35\linewidth,
trim=140px 90px 75px 60px,clip]
{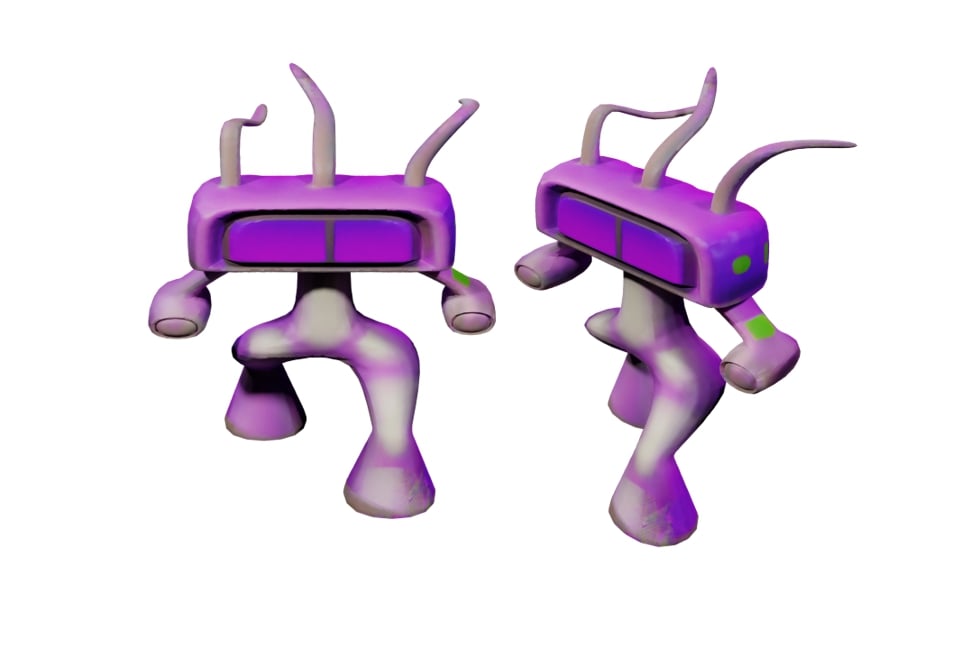}&
\includegraphics[width=.35\linewidth,
trim=140px 90px 75px 60px,clip]
{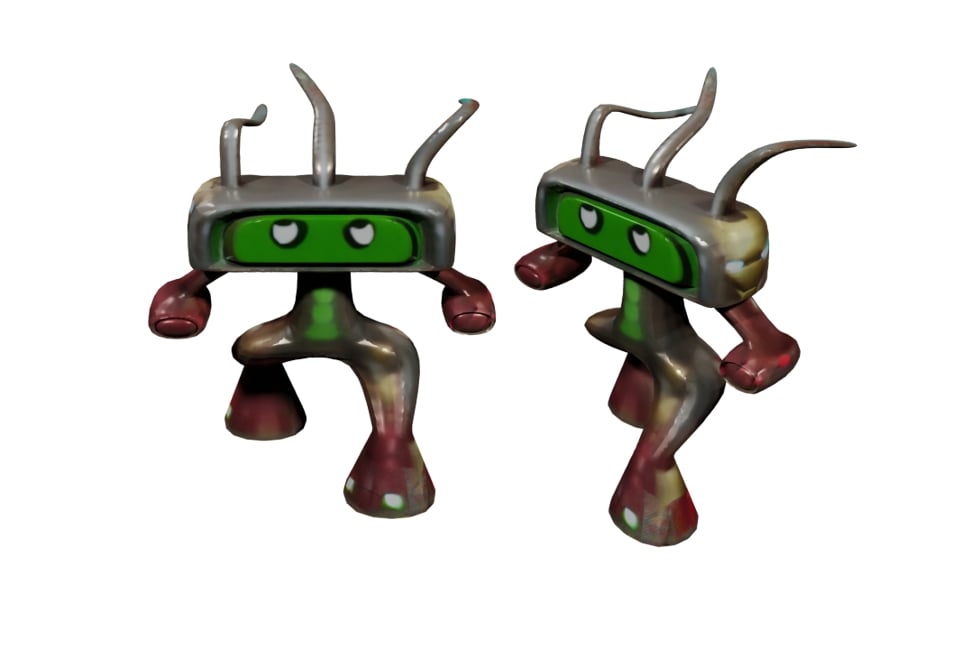}\\
\TWORCell{Wired alien with}{purple skin, TV head}&
\TWORCell{Cartoon alien with}{iron man skin, TV head}\\
\includegraphics[width=.35\linewidth,
trim=170px 130px 170px 120px,clip]
{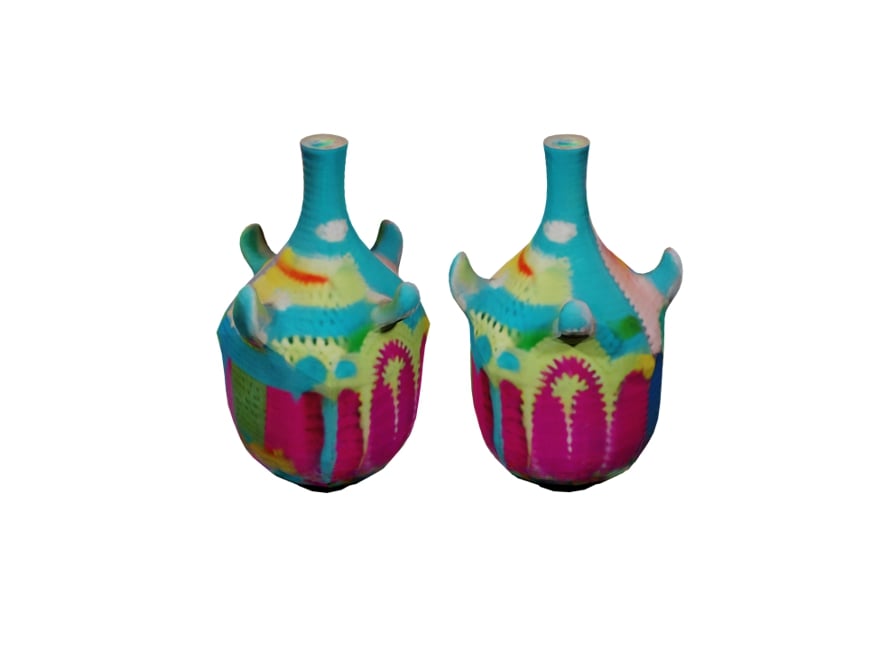}&
\includegraphics[width=.35\linewidth,
trim=170px 130px 170px 120px,clip]
{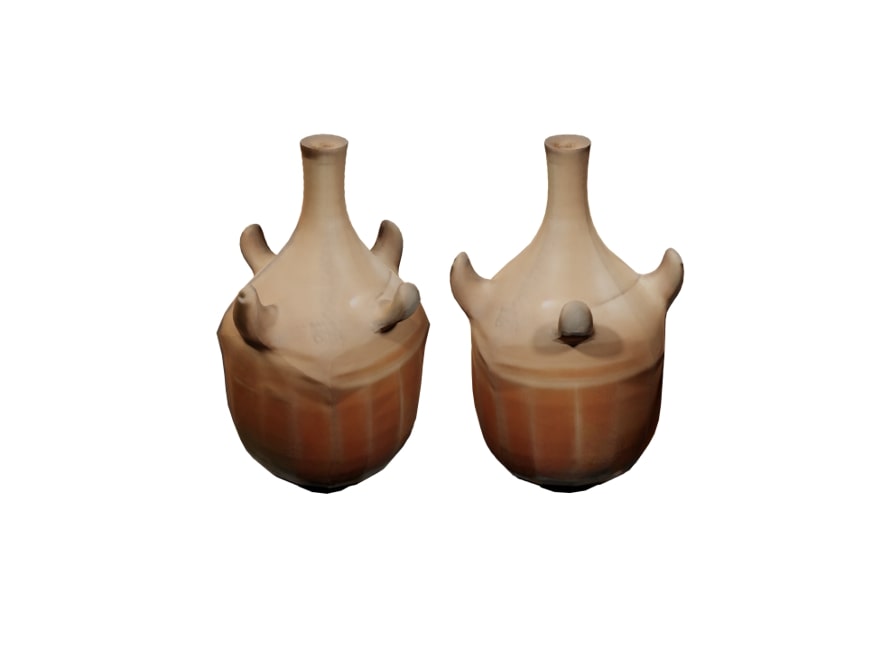}\\
A colorful crochet vase&
Ancient earthware pot\\
\end{tabular}
\caption{\label{fig:diffPrompts} Generated textures from different prompts.}
\vspace{-6mm}
\end{figure}

\section{Evaluation}
All experiments are run on a machine with RTX 3090 GPU, an Intel Core i7-11600 CPU, and 16GB RAM. The total time for generating one texture is around $25$ to $30$ minutes. 
The pipeline utilizes the pre-trained LDM, Stable-Diffusion-2-Depth, which takes a depth map as the conditional input, generates a 2D latent code with $64\times64\times4$ resolution, and decodes it into an image with a resolution of $512\times512\times3$ in color space.
To ensure a fair evaluation of our method, we adopt an identical configuration to render all mesh models in experiments. Initially, all meshes are normalized within a bounding box of unit length, and the UV are automatically mapped by XAtlas~\cite{xatlas}. Subsequently, we sample 8 fixed camera positions on a sphere with 1.5 radius, using a 45-degree field of view, 30-degree pitch, and evenly spaced yaw angles ranging from 0 to 315 degrees. Since illumination can disrupt the normal distribution during the reversed diffusion process, we opt not to apply lighting and shading during rendering.
During the reversed diffusion process, we run $35$ denoising steps. In each step, we optimize $\{\bar{z}_{0,t}^i\}$ using $20$ iterations of AdamW optimizer with learning rate of 0.01. Afterward, we synthesis the final color texture from all camera views. To accomplish this, we employ an optimization to match the color of texels with the corresponding colors in image views, using $500$ iterations of SGD. The resolution of final texture is $1024\times1024\times3$. 
We show the results from our system in Figure~\ref{fig:teaser}. In Figure~\ref{fig:diffPrompts}, our method works well with different text prompts on the same input mesh. We also display more generation results in the supplementary material.

We focus the comparison on TEXTure~\cite{richardson2023TEXTure}, TexFusion~\cite{cao2023texfusion}, Meshy, and Fantasia3D~\cite{Chen_2023_Fantasia3D}. TEXTure performs the denoising process sequentially for each view and paints the texture in RGB space directly. The LDM they used is same with our backbone. TexFusion tightly couples the denoising process and latent-space fusion. Meshy is an industrial tool for texture generation. 
Fantasia3D is a SDS-based method, generating both models and textures. It also supports the synthesis of textures from a geometry input separately.
Our evaluation is performed over $64$ high-quality mesh models selected from 3D model datasets and multiple online resources and we use $182$ pairs of meshes and text prompts for the experiments.

\paragraph{Qualitative Comparison} We displayed the texture generation results and the comparison of other four methods in Figure~\ref{fig:Qualitative}.
In this experiment, we used the official implementation provided by TEXTure, Meshy and Fantasia3D. As the code for TexFusion is not released, we made a local implementation for this comparison. From the results, we found the high quality of our generated textures were consistent across all these models, while other texture painting methods had different problems with multi-view consistency. 
Additionally, the results from Fantasia3D sometimes exhibit over-exposure and over-saturation in color.  
Of the four techniques, we observe that the generation process in Fantasia3D exhibits issues with unnatural color generation.
The TEXTure exhibits relatively worse result, exhibiting noisy texture patches. The results generated by Meshy have a high variation, giving stunning results on some models but low-quality textures on others. The results by TexFusion have the highest quality of the four while being consistently better than either TEXTure or Meshy. However, by detailed observation, our method achieves better quality and multi-view consistency.
In addition to the comparison with our local implementation of TexFusion, we also reproduced several results with the same input meshes and text prompts as~\cite{cao2023texfusion}. We supplemented the side-by-side comparison in Figure~\ref{fig:TexFusionQualitative}, where we adjusted our camera views and lighting conditions as much as possible for a fair comparison.

\paragraph{Quantitative Comparison} We further compared the quality of textures using the FID~\cite{heusel2018gans} metric. This is a learned metric to measure the similarity between two datasets of images, in terms of visual quality and human preference. Since our method and TEXTure use SD2-depth as backbone, and Fantasia uses SD2, we propose to use the images generated by SD2-depth as the groundtruth dataset. Specifically, after generating our texture, we rendered the model under the given set of 8 camera views to generate our dataset. We then queried SD2-depth conditioned on the depth map from each camera view, along with the text prompt, to generate the groundtruth dataset. The resulting FID metrics for the four methods are summarized in Table~\ref{table:FID}. The FID scores reflect the same observations we made in our qualitative comparison. Note that this comparison can be unfair for Meshy because we do not know their backbone LDM model. But since SD2-depth is trained using large-scale internet dataset, we assume the comparison is reasonable. 
Besides, we also measure the average time consumption of all methods during in the texture generation. The measurements are summarized in Table~\ref{table:time}.

\paragraph{Ablation Study} Finally, we show that each and every component of our method is necessary. 
First, we highlighted the benefits of working with color space. In the experiment, we directly blended $\{\hat{z}_{0,t}^i\}$ into a latent texture $I_{z,0,t}$ and then rendered it under corresponding view to yield $\{\bar{z}_{0,t}^i\}$, keeping other components intact. 
As shown in Figure~\ref{fig:ablation} (left), 
this variant resulted in inconsistent and blurry textures. 
Next, we show that our method could only work with DDIM because it predicted a noiseless latent $\{\hat{z}_{0,t}^i\}$. 
For comparison, we combined our method with DDPM~\cite{ho2020DDPM} and applied our color fusion on the noisy latent $\{z_{t}^i\}$, i.e., we decoded $\{z_t^i\}$ to $\{x_t^i\}$, blended it into $I_{x,0,t}$ and then optimized to yield $\{\bar{z}_{0,t}^i\}$. 
As shown in Figure~\ref{fig:ablation} (middle), 
this variant resulted in meaningless textures. This was due to the inconsistent noise component across views. 
In Equation~\ref{eq:FuseZ}, we employs an optimization to achieve the conversion from a color image to a latent code. A more direct approach is to utilize the VAE encoder of LDM built-in. However, frequent encoding of the image during generation can slightly decreases the quality of the final texture in some cases, despite significantly faster computation. In our quantitative experiment, we also measure the FID score and time consumption of our method with a VAE encoding, and the results are presented in Table~\ref{table:FID} and Table~\ref{table:time}. We illustrate a case that encounters encoding issues in Figure~\ref{fig:ablation} (right).

\begin{table}[h]
\centering
\caption{\label{table:FID}FID score of the five methods and our method with direct encoding(DE) under comparison.}
\begin{tabular}{cccccc}
\toprule
TEXTure & Meshy & Fantasia3D & TexFusion & Ours & Ours(DE) \\
\midrule
$88.16$ & $70.49$ & $57.25$ & $41.55$ & $38.21$ & $43.40$\\
\bottomrule
\end{tabular}
\end{table}

\vspace{-6mm}

\begin{table}[h]
\centering
\caption{\label{table:time}The time consumption of the five methods and our method with direct encoding(DE) under comparison in minutes.}
\begin{tabular}{cccccc}
\toprule
TEXTure & Meshy & Fantasia3D & TexFusion & Ours & Ours(DE) \\
\midrule
$5.5$ & N/A & $24.1$ & $4.2$ & $27.3$ & $4.8$\\
\bottomrule
\end{tabular}

\vspace{-3mm}
\end{table}
\section{Conclusion}
We present TexPainter, a new approach to generate semantic textures for arbitrary, given 3D models, utilizing a pre-trained LDM image generator. Although this problem has been studied by prior works, we show that our method achieves better texture quality, via a new approach to enforce multi-view consistency. Our main idea is to utilize the predicted noiseless state in DDIM to perform multi-view fusion in the color space and then update the noiseless state using an optimization. Our approach further eliminates the assumption on sequential dependency between views. Through extensive experiments and comparisons, we confirm that our method delivers consistently high quality textures. We further highlight the potential of our method as a general method for enforcing constraints between multiple diffusion process. 

However, we still notice several shortcomings that worth further exploration. First, our method still uses a fixed set of camera views, while prior work~\cite{chen2023text2tex} proposed an algorithm to select a dynamic set of views for better covering of the model. We emphasize that it is non-trivial to combine a dynamic set of views with our method, because images from all the views must be present and fused together during each denoising step, making incremental view selection difficult. 
Further, our method inherits the limitations of prior works, i.e., the texture quality is limited by the pre-trained LDM and the sharp features cannot be generated well. 
Due to the lack of specific 3D directional constrains on the 2D LDM, it can sometimes erroneously generate multiple faces for a human model. Additionally, inadequate optimization and the weighted average operation potentially lead to blurry regions and smooth edges. The problem can be resolved in multiple ways in the future, including using a better pre-trained model, a texture up-sampling approach such as~\cite{sajjadi2017enhancenet}, or finding an optimal method for texture synthesis while maintaining consistency in color space. Besides, utilizing a fine-tune diffusion model~\cite{zhang2023controlnet} could enhance the overall quality of generated texture or the alignment with specific requirements.
In addition, our method incurs higher computational cost than prior works~\cite{cao2023texfusion} due to the repeated optimization. A lossless encoder would circumvent the time-intensive optimization in Equation~\ref{eq:FuseZ}, thereby significantly enhancing the performance of our method.

\begin{acks}
This work has been supported by the Natural Science Foundation of Jiangsu Province Project award BK20232008, BK20211159, Jiangsu Key Research and Development Plan under Grant award BE2023023, the Joint Fund Project award 8091B042206 and the Fundamental Research Funds for the Central Universities. It is also partially supported by the Innovation and Technology Commission of the HKSAR Government under the InnoHK initiative.
\end{acks}

\bibliographystyle{ACM-Reference-Format}
\bibliography{reference.bib}

\newpage
\appendix
\begin{figure*}[ht]
\centering
\includegraphics[width=1.0\linewidth,trim=130px 60px 20px 60px,clip]{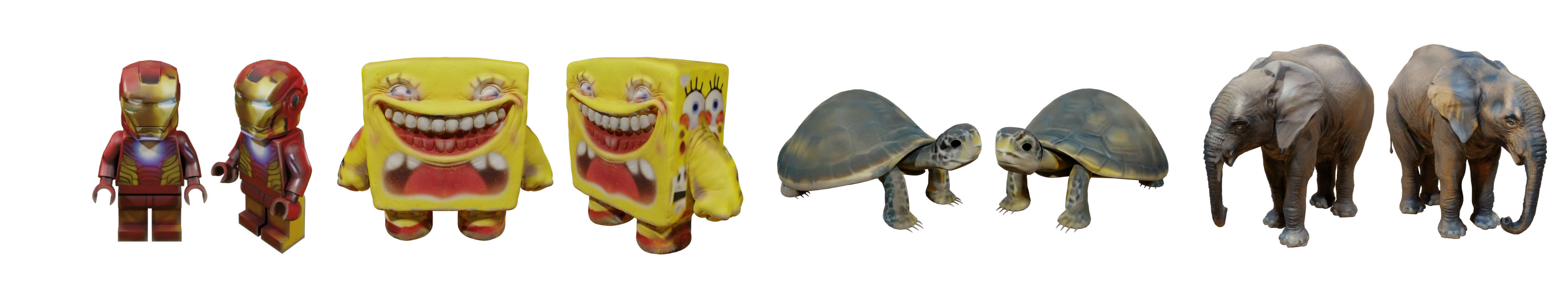}
\put(-500,-10){a LEGO iron man}
\put(-405,-10){Yellow SpongeBob SquarePants}
\put(-415,-20){laughing heartily, looking frightening}
\put(-215,-10){A turtle}
\put(-85,-10){African elephant}
\caption{\label{fig:MoreResults}More results generated using our method.}
\end{figure*}

\begin{figure*}[ht]
\centering
\includegraphics[width=1.0\linewidth]{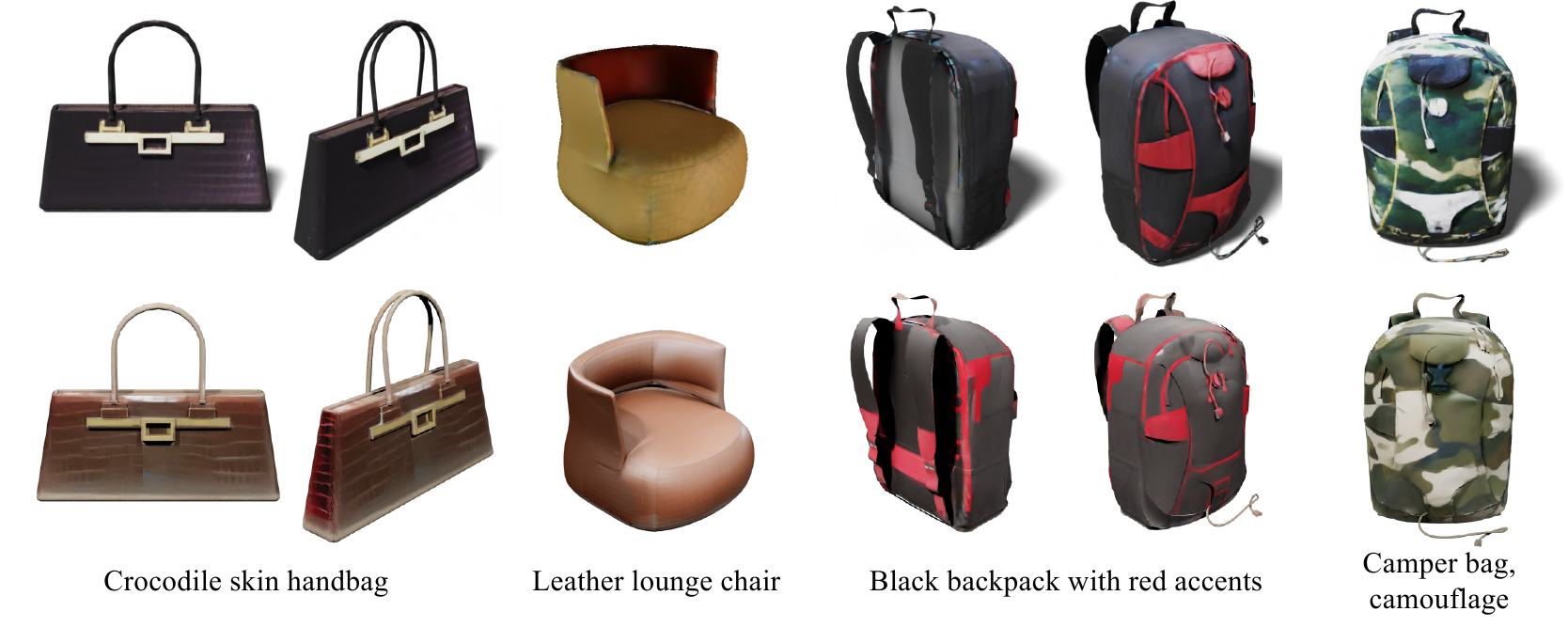}
\caption{\label{fig:TexFusionQualitative}Comparison between TexFusion and our method. We generate textures on the same meshes with the same prompts used in TexFusion (top) and render our results using similar views and lighting conditions (bottom).}
\end{figure*}

\begin{figure*}[ht]
\centering
\setlength{\tabcolsep}{0px}
\begin{tabular}{cccccc}
\includegraphics[width=.1\linewidth,trim=70px 20px 100px 20px,clip]{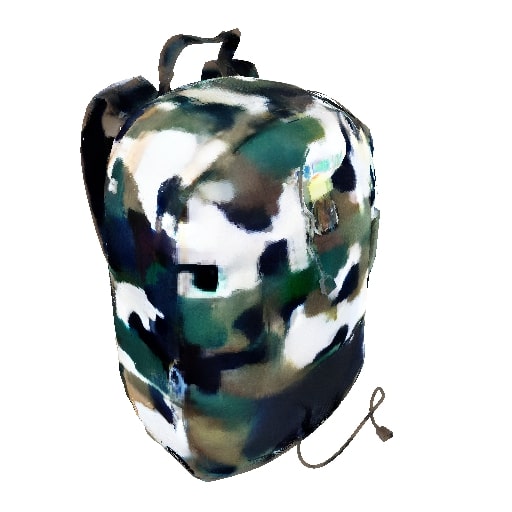}&
\includegraphics[width=.1\linewidth,trim=70px 20px 100px 20px,clip]{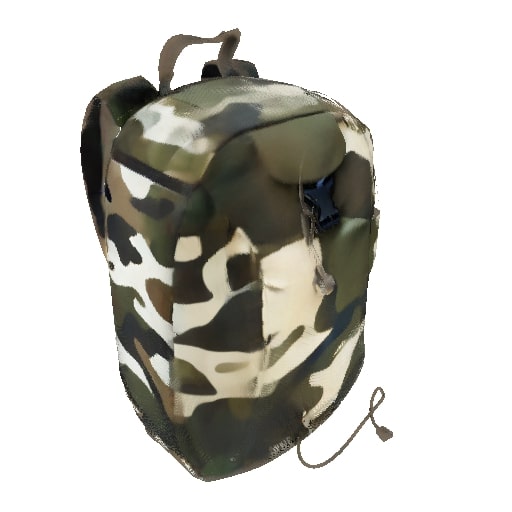}&
\includegraphics[width=.16\linewidth,trim=50px 80px 90px 100px,clip]{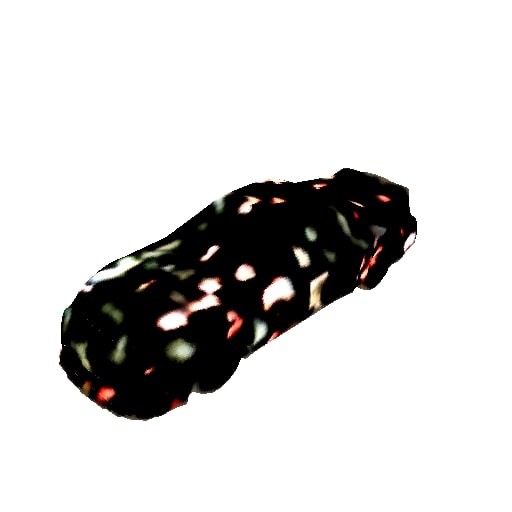}&
\includegraphics[width=.16\linewidth,trim=50px 80px 90px 100px,clip]{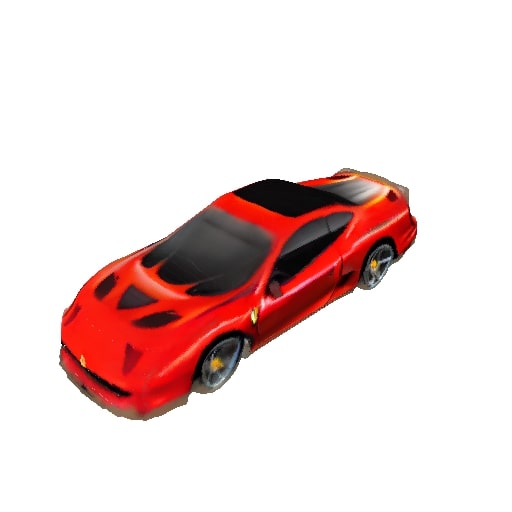}&
\includegraphics[width=.20\linewidth,trim=0px 0px 0px 0px,clip]{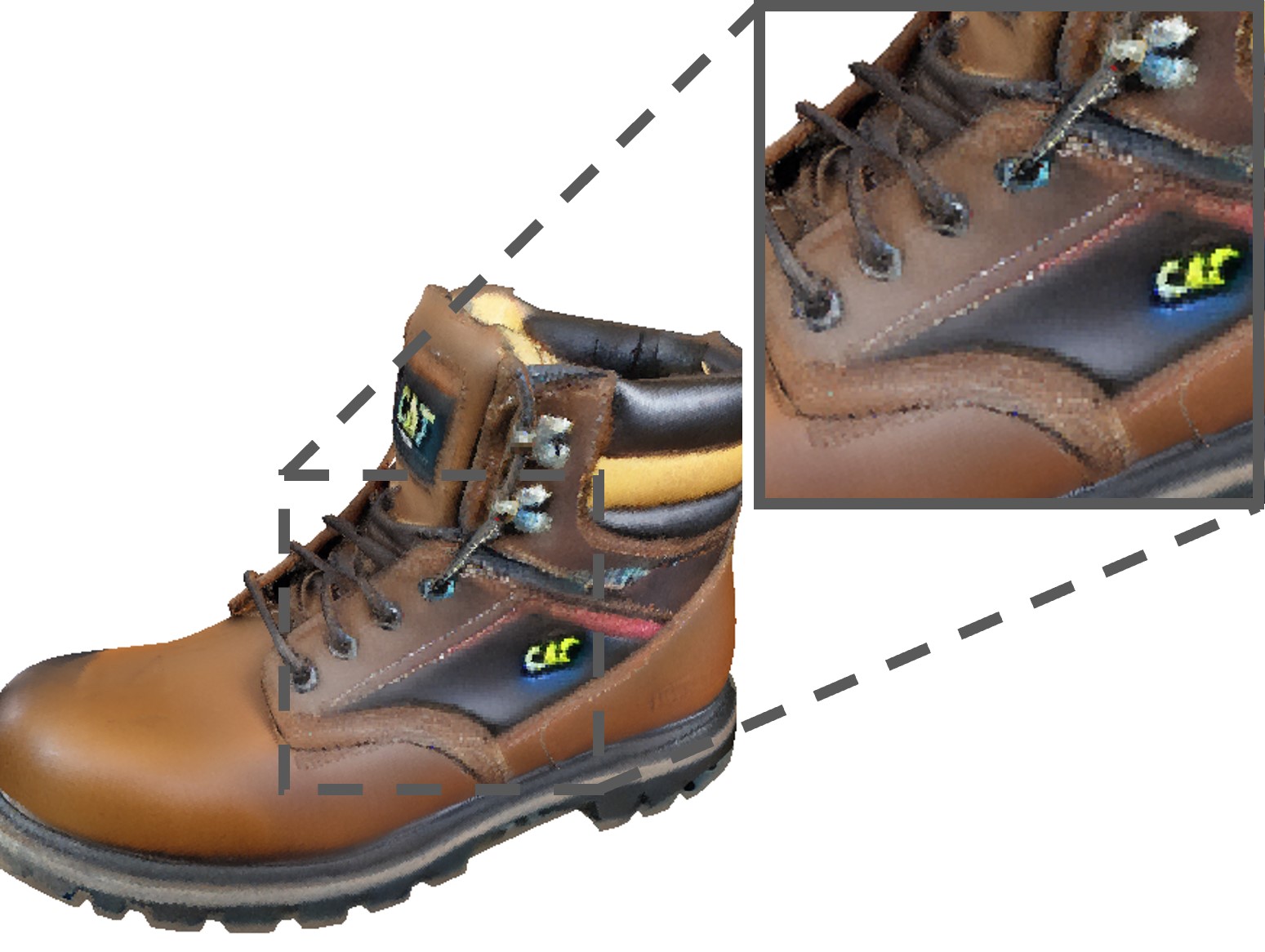}&
\includegraphics[width=.20\linewidth,trim=0px 0px 0px 0px,clip]{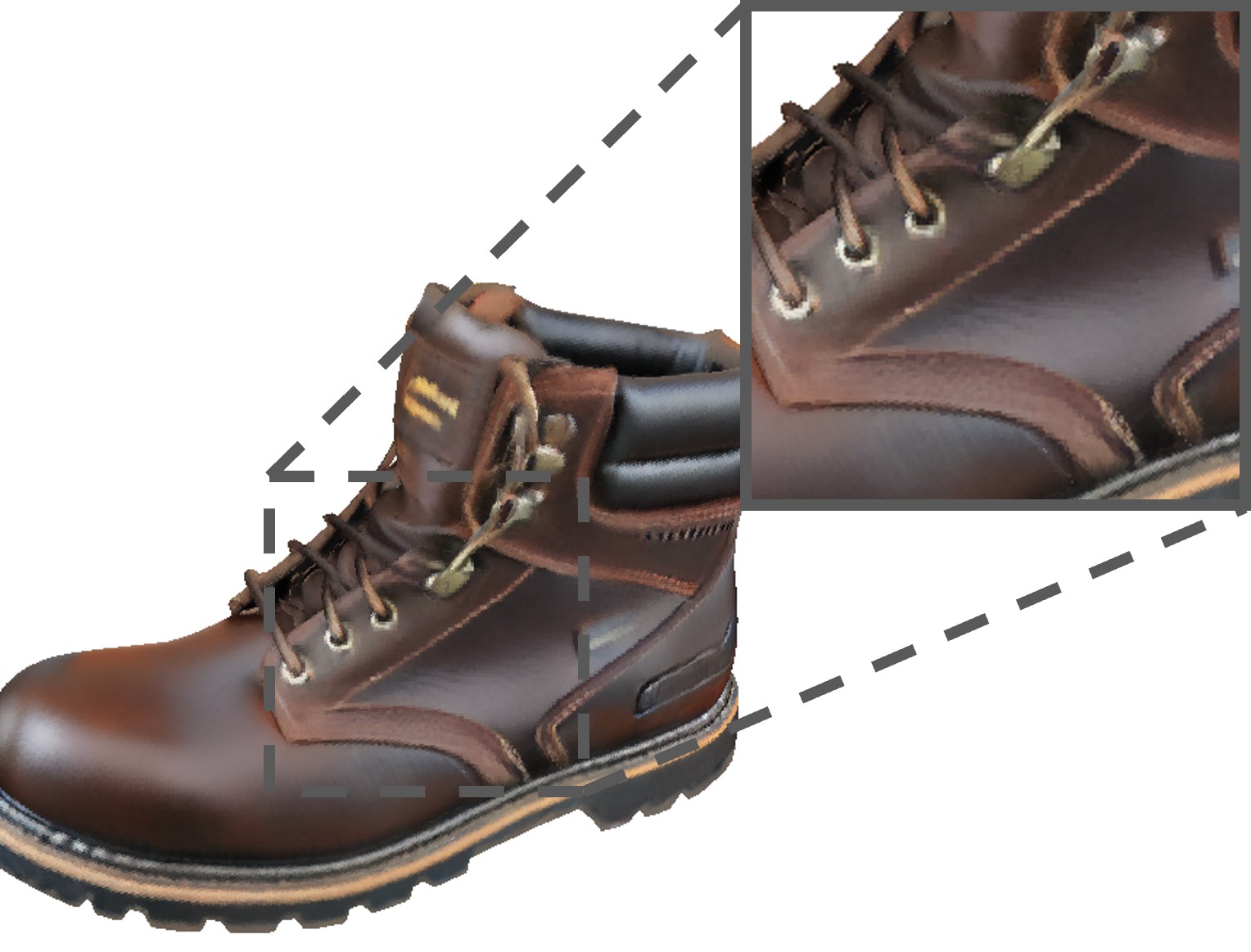}\\
Direct latent blending &
Ours &
Fusing noisy images &
Ours&
Direct encoding &
Ours\\
\end{tabular}
\caption{\label{fig:ablation}
Ablation study of our proposed method. Left: Compared with our method that works in color space, directly blending the latents would result in inconsistent and blurry textures. Middle: If we blend the noisy images predicted using DDPM instead of using the noiseless images predicted using DDIM, the resulting textures could be meaningless. Right: The repeated direct encoding decreases the quality of the final texture.}
\end{figure*}

\begin{figure*}[ht]
\centering
\setlength{\tabcolsep}{0px}
\begin{tabular}{ccc}
\includegraphics[width=.30\linewidth,trim=0px 0px 0px 0px,clip]{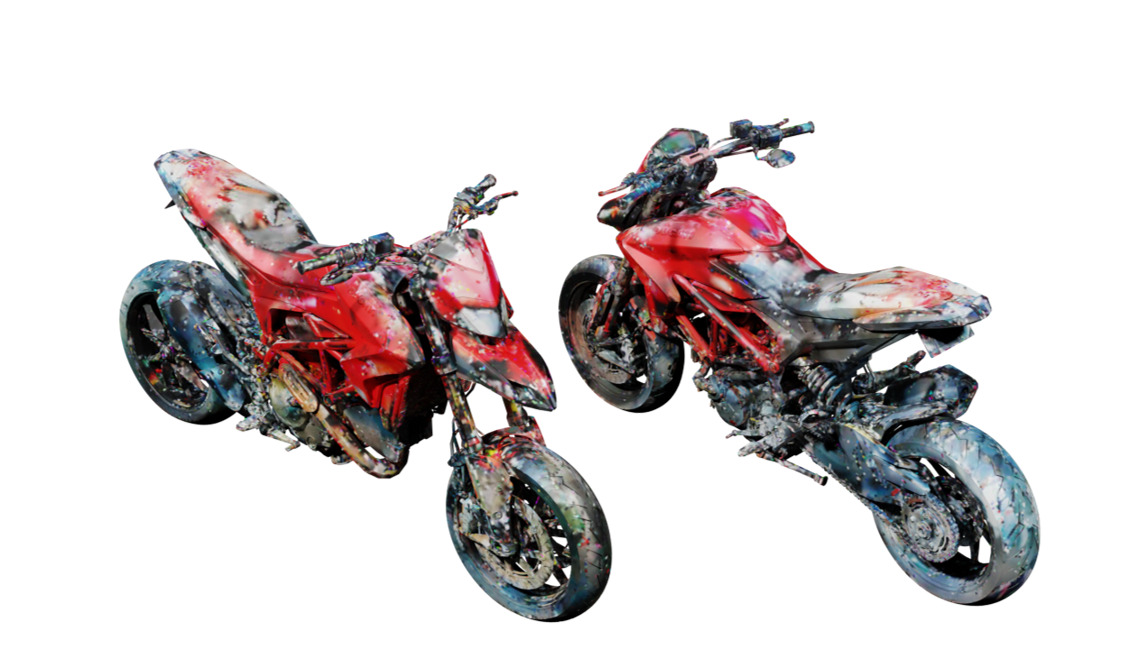}
\put(-150,100){TEXTure}&
\includegraphics[width=.30\linewidth,trim=0px 0px 0px 0px,clip]{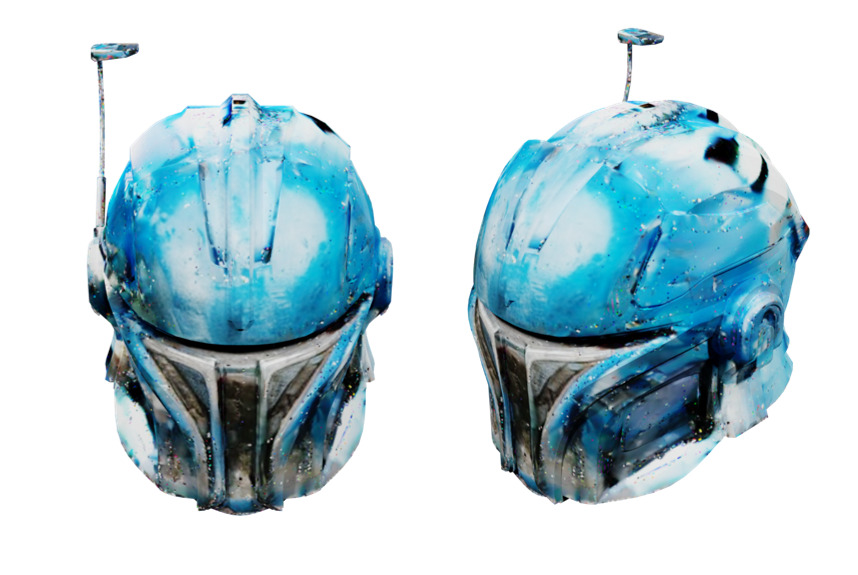}&
\includegraphics[width=.30\linewidth,trim=0px 0 0px 0,clip]{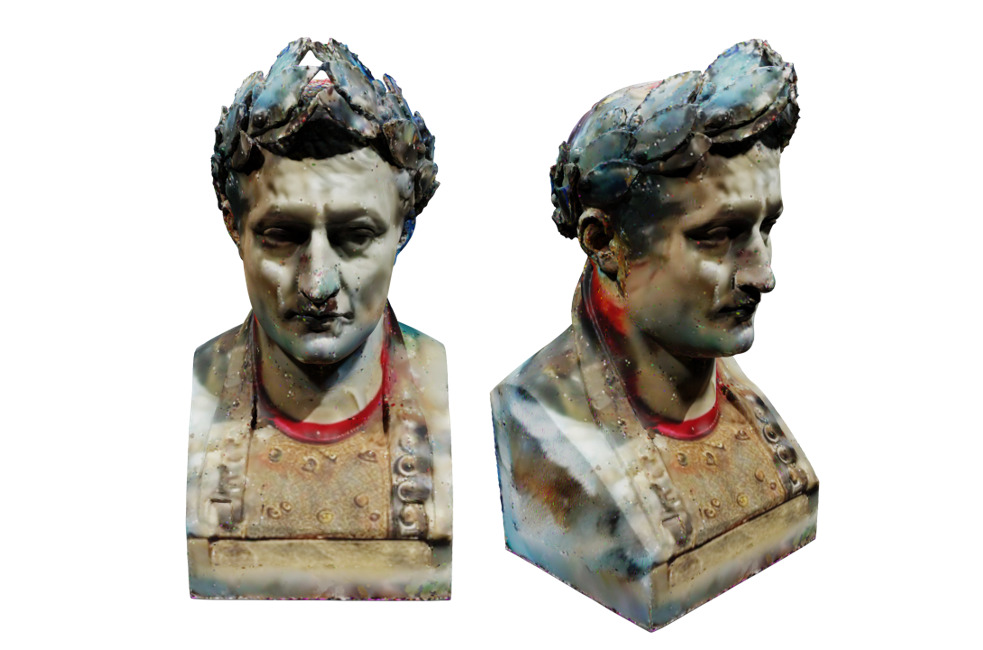}\\
\includegraphics[width=.30\linewidth,trim=0px 0px 0px 0px,clip]{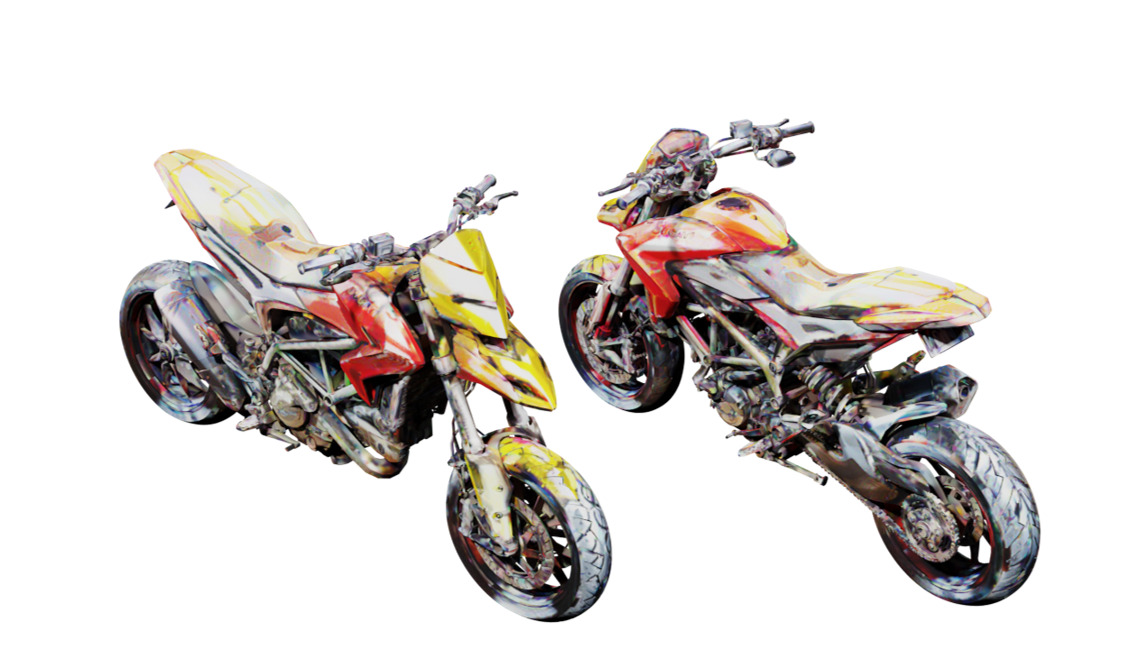}
\put(-150,100){Meshy}&
\includegraphics[width=.30\linewidth,trim=0px 0px 0px 0px,clip]{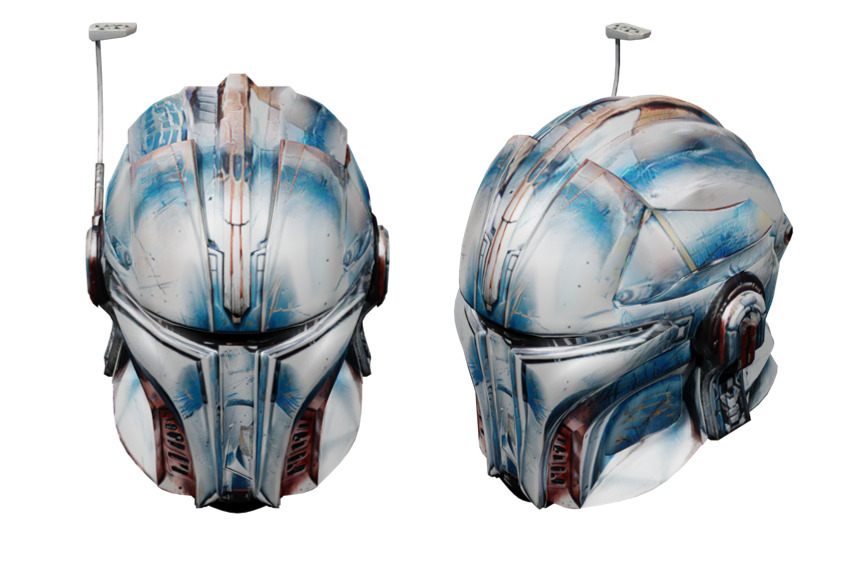}&
\includegraphics[width=.30\linewidth,trim=0px 0 0px 0,clip]{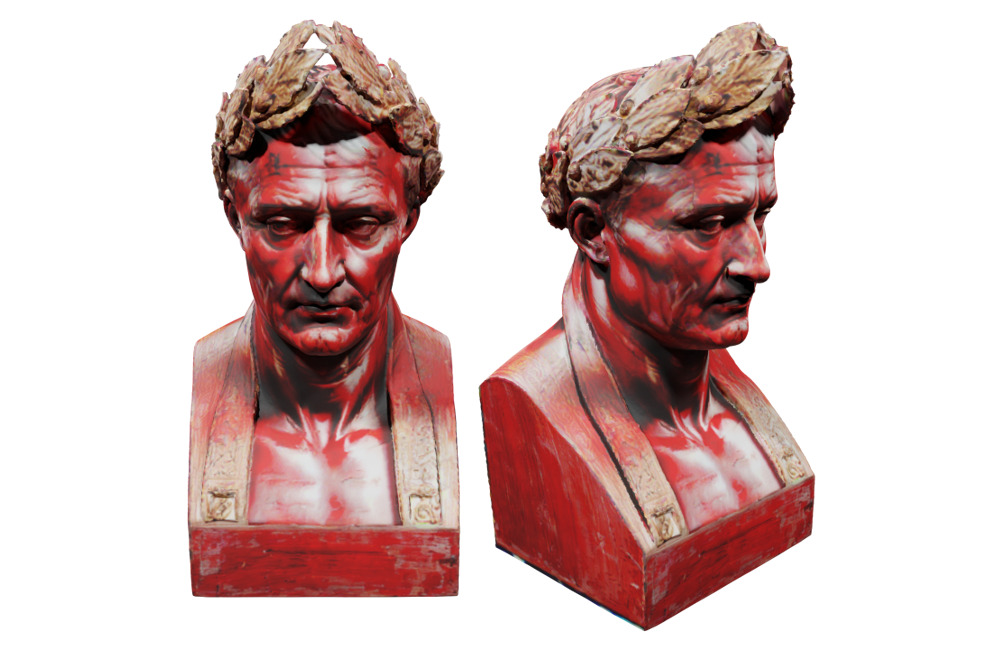}\\
\includegraphics[width=.30\linewidth,trim=0px 0px 0px 0px,clip]{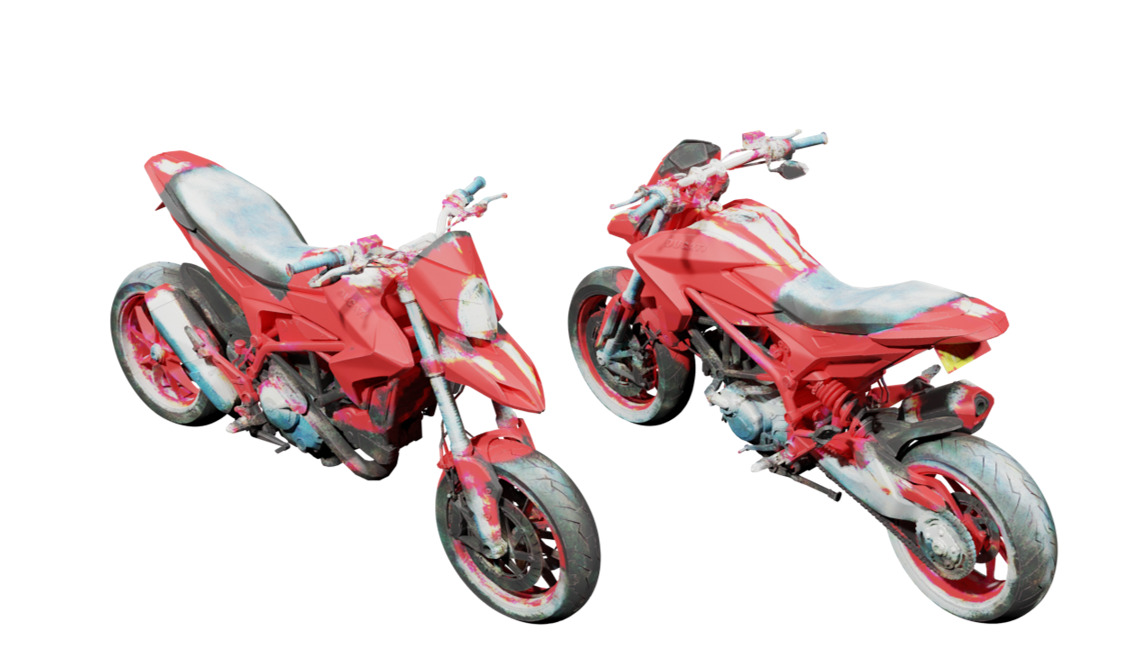}
\put(-150,100){Fantasia3D}&
\includegraphics[width=.30\linewidth,trim=0px 0px 0px 0px,clip]{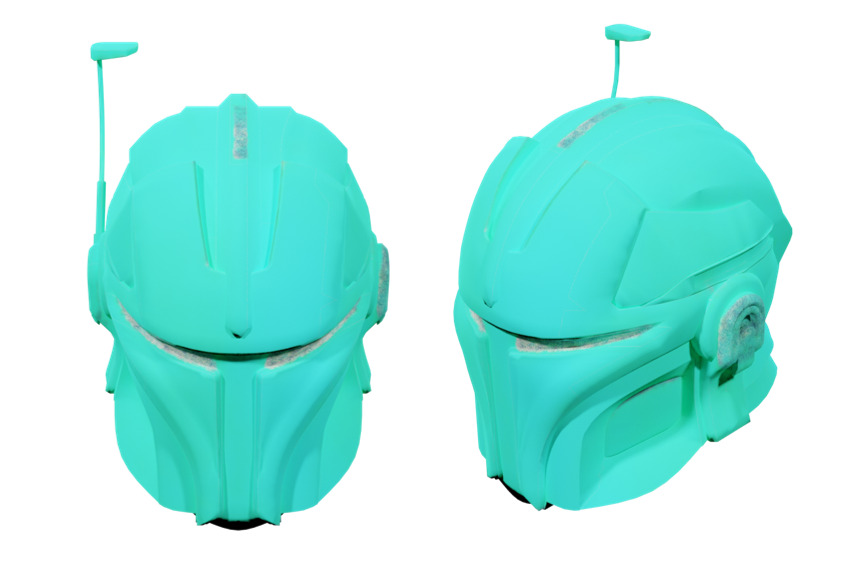}&
\includegraphics[width=.30\linewidth,trim=0px 0 0px 0,clip]{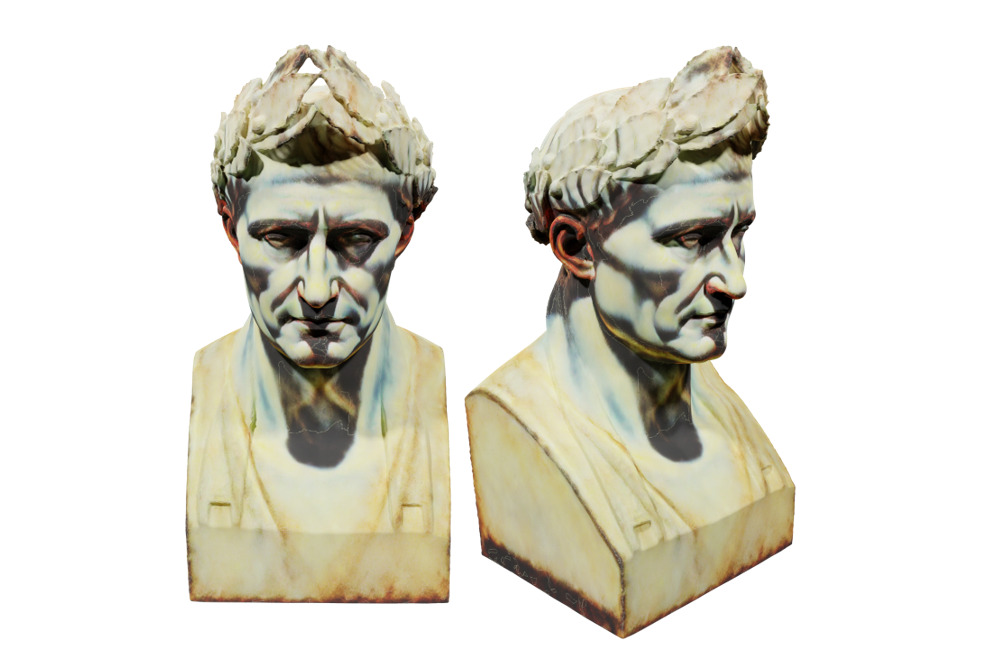}\\
\includegraphics[width=.30\linewidth,trim=0px 0px 0px 0px,clip]{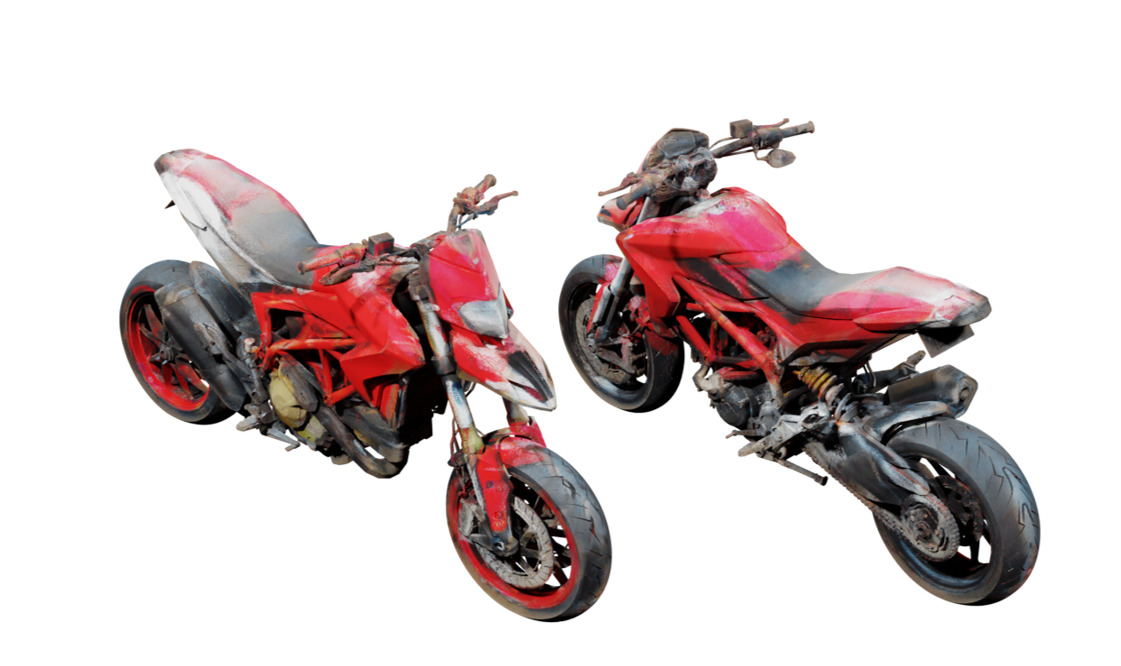}
\put(-150,100){TexFusion}&
\includegraphics[width=.30\linewidth,trim=0px 0px 0px 0px,clip]{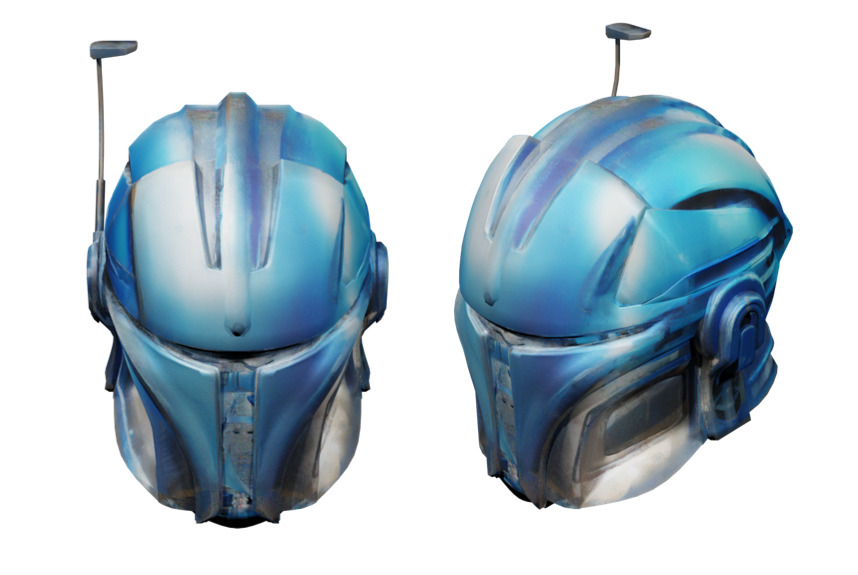}&
\includegraphics[width=.30\linewidth,trim=0px 0 0px 0,clip]{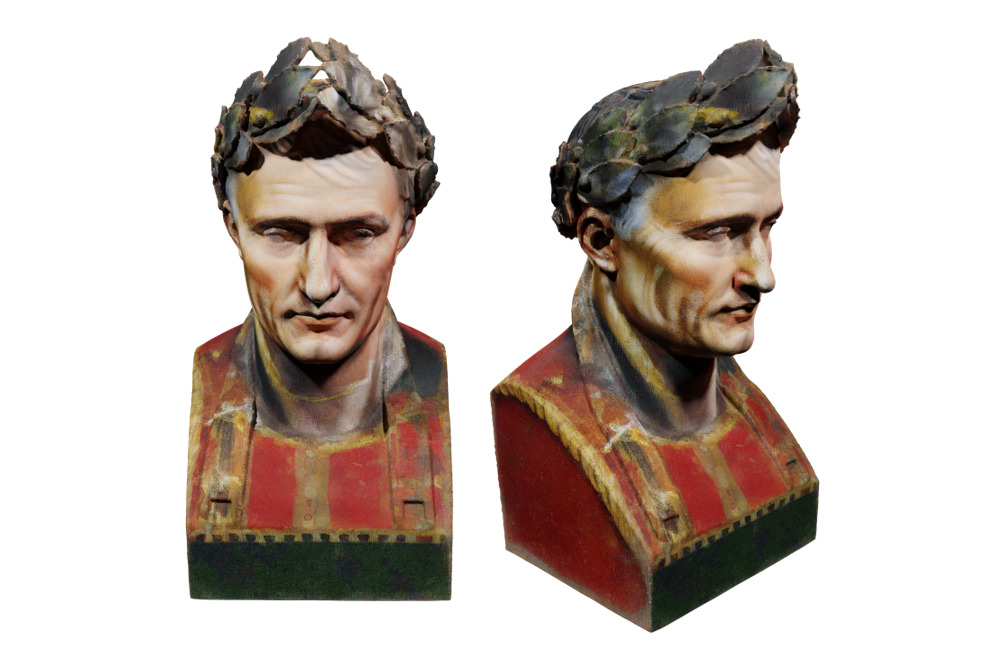}\\
\includegraphics[width=.30\linewidth,trim=0px 0px 0px 0px,clip]{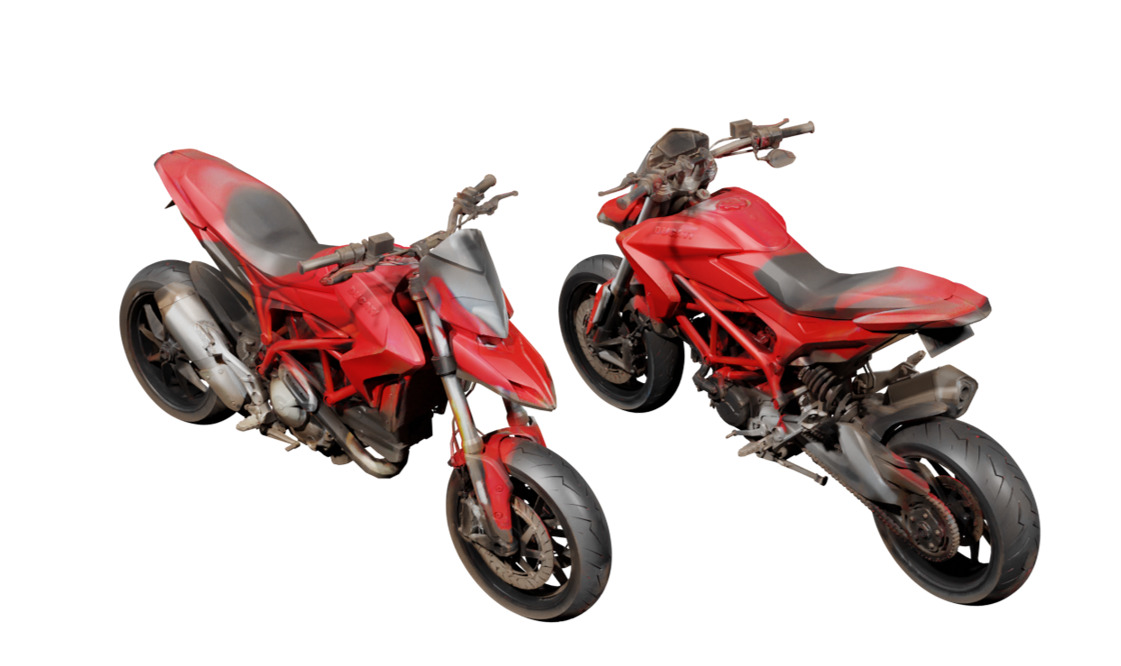}
\put(-150,100){Ours}&
\includegraphics[width=.30\linewidth,trim=0px 0px 0px 0px,clip]{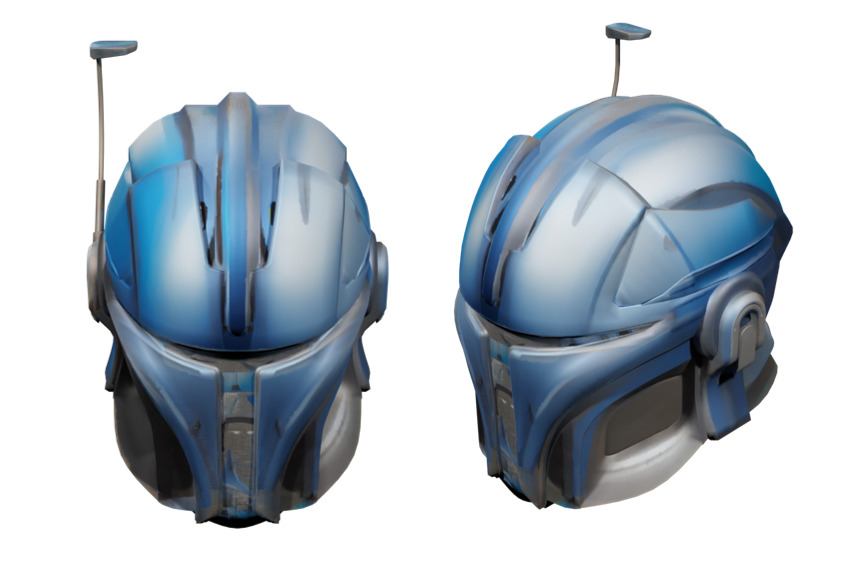}&
\includegraphics[width=.30\linewidth,trim=0px 0 0px 0,clip]{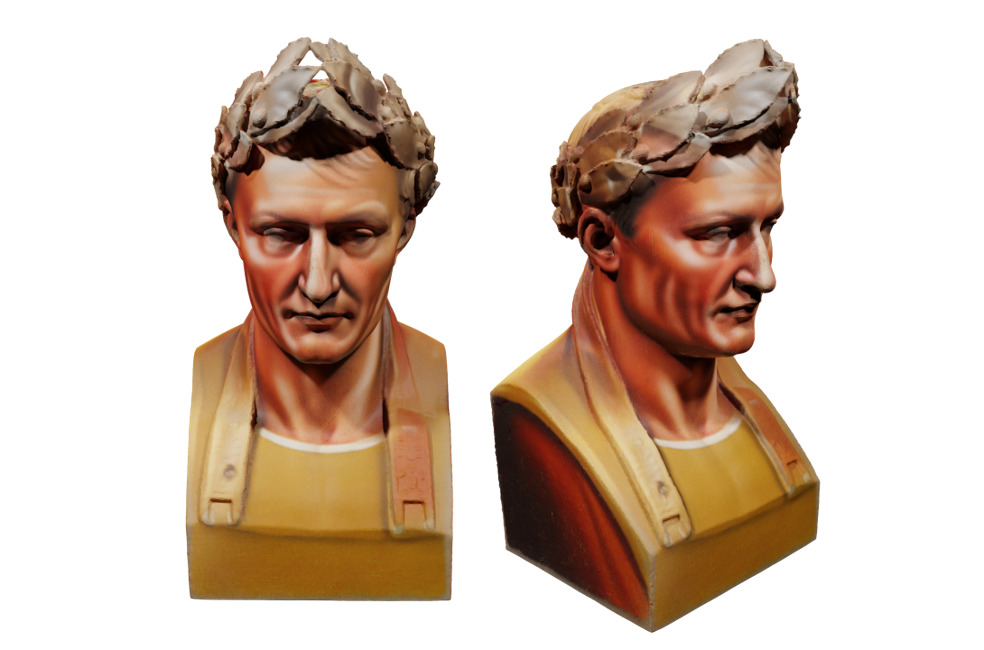}\\
motorbike, Ducati Hypermotard 939&
a blue helmet, sci-fi movie&
Julius Caesar, oil painting, full color\\
\end{tabular}
\caption{\label{fig:Qualitative}Comparison of textures generated by TEXTure, Meshy, Fantasia3D, our implementation of TexFusion, and our method.}
\end{figure*}

\newpage
\end{document}